\documentclass[11pt]{article}

\usepackage[final]{acl}

\usepackage{times}
\usepackage{latexsym}

\usepackage[T1]{fontenc}

\usepackage[utf8]{inputenc}

\usepackage{microtype}

\usepackage{inconsolata}

\usepackage{graphicx}

\title{Zipping the Thought: When and How Compressed Reasoning Data \\ Works in LLM Post-Training}

\author{
  \textbf{Kohsei Matsutani$^\dagger$, Gouki Minegishi, Takeshi Kojima} \\
  \textbf{Yusuke Iwasawa, Yutaka Matsuo} \\
  The University of Tokyo \\
  $^\dagger$\texttt{kohsei.matsutani@weblab.t.u-tokyo.ac.jp}
}

\usepackage[utf8]{inputenc} %
\usepackage[T1]{fontenc}    %
\usepackage{hyperref}       %
\usepackage{url}            %
\usepackage{booktabs}       %
\usepackage{amsfonts}       %
\usepackage{nicefrac}       %
\usepackage{microtype}      %
\usepackage{xcolor} 
\usepackage{amsmath,amsfonts,bm}
\usepackage{cleveref} 
\usepackage{xurl}
\usepackage{amsmath,amssymb,amsthm,mathtools}
\usepackage[most]{tcolorbox}
\tcbuselibrary{listings, skins, breakable}
\usepackage{listings}
\usepackage{tikz}
\usetikzlibrary{calc}
\usepackage{graphicx}
\usepackage{subcaption}
\usepackage{multirow}
\usepackage{tabularx}
\definecolor{HeaderGray}{gray}{0.92}
\usepackage{float}
\usepackage{colortbl}
\usepackage{wrapfig}
\usepackage{enumitem}
\usepackage{adjustbox}
\usepackage{breqn}
\usepackage{hyperref}
\usepackage{cleveref}

\usepackage[most]{tcolorbox}
\usepackage{xcolor}
\usepackage{lipsum}
\usepackage{hyperref}

\definecolor{opcolor}{HTML}{B84A4A}
\definecolor{valcolor}{HTML}{3B6EA8}

\newtcolorbox{takeaway}{
  enhanced,
  breakable,
  frame hidden,
  colback=gray!15!white,
  arc=2mm,
  left=2mm,
  right=2mm,
  top=1.5mm,
  bottom=1.5mm
}

\newtcblisting{promptbox}[2][]{%
  enhanced,
  breakable,
  colback=white,
  colframe=gray!75!black,
  title={#2},
  listing only,
  listing engine=listings,
  listing options={
    basicstyle=\ttfamily\small,
    breaklines=true,
    breakatwhitespace=false,
    columns=fullflexible,
    keepspaces=true,
    showstringspaces=false,
    language={},
    keywordstyle=,
    commentstyle=,
    stringstyle=,
  },
  #1
}

\begin{document}
\maketitle
\begin{abstract}
Large language models (LLMs) can now solve complex problems through long chain-of-thought (CoT) reasoning, but the trade-off between performance and token cost remains a central challenge. To address this issue, supervised fine-tuning (SFT) often uses compressed reasoning data, where CoT traces are shortened into compact forms. However, the effect of such compressed reasoning data on post-training remains poorly understood. In this paper, we propose a taxonomy of CoT consisting of Explicit CoT, which outputs all operations without aggregation, Composed CoT, which combines multiple operations into a single step, and Implicit CoT, which omits intermediate operations. As a controlled mechanistic study, we construct a synthetic compositional reasoning task that allows controlled variation of difficulty, compression granularity, and data size, and conducted a comprehensive set of experiments across different model families and sizes. Notably, we find that (i) coarser CoT requires more SFT data, (ii) compared with Explicit CoT, Composed CoT and Implicit CoT benefit more from data scaling, while Composed CoT benefits from data repetition and Implicit CoT tends to lead to memorization, (iii) unlike SFT, subsequent reinforcement learning (RL) with verifiable rewards (RLVR) decomposes compressed steps learned during SFT, and (iv) unidirectional CoT ordering shows stronger generalization, on longer sequential tasks. Our findings provide implications for CoT design under data resource constraints and offer important insights into the mechanisms of SFT and RL in LLM post-training.
\end{abstract}

\section{Introduction}

Algorithmic tasks can be formulated as the sequential composition of simpler subproblems \citep{bellman1957dynamic,newell1972human}. In large language models (LLMs), such compositional structure is often realized through chain-of-thought (CoT) reasoning \citep{wei2022chain,kojima2022large}, which externalizes intermediate reasoning steps as tokens before producing the final answer. This capability is often enhanced during supervised fine-tuning (SFT) and reinforcement learning (RL) with verifiable rewards (RLVR) at post-training \citep{openai2024jaech,deepseekai2025deepseekr1,lambert2025tulu}.
\begin{figure}[!t]
  \centering
  \includegraphics[width=0.95\columnwidth]{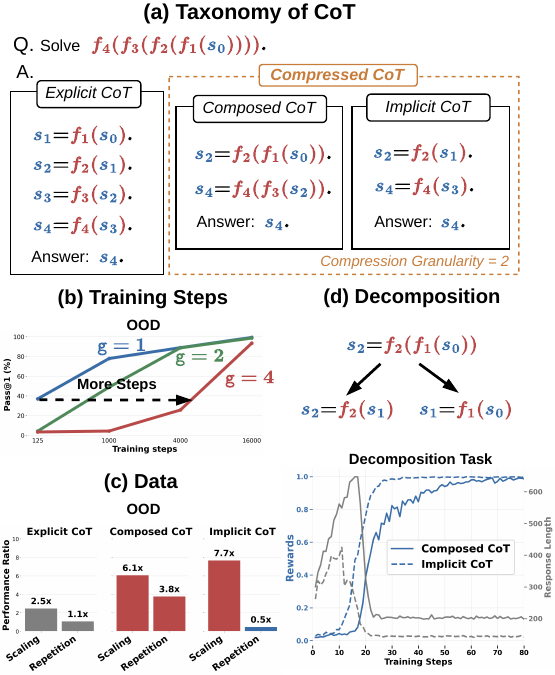}
  \caption{\textbf{(a) Taxonomy of CoT.} Example of an $\texttt{op}=4$ task. $\textcolor{opcolor}{f_i}$ denotes an operation and $\textcolor{valcolor}{s_i}$ denotes a value. See \Cref{fig:data_main} for task descriptions. \textbf{(b) CoT Granularity in SFT and Training Steps.} Coarser-grained CoT SFT requires more training steps. \textbf{(c) Data Scaling vs. Data Repetition.} Compressed CoT benefits substantially from data scaling. Composed CoT benefits from data repetition, whereas Implicit CoT is adversely affected by it. \textbf{(d) Decomposition of Steps by RLVR.} On-policy exploration in RLVR can decompose compressed reasoning steps. Results in (b), (c), and (d) are from Qwen2.5-3B.}
  \label{fig:main}
\end{figure}

Despite the remarkable success of CoT reasoning, the widespread deployment of LLM agents has escalated token costs \citep{bai2026how}, motivating efforts to compress reasoning traces without compromising performance. 

While prior studies have proposed CoT compression methods, shortening long CoT into a compact form, such as SFT or self-distillation \citep{huang2026reasoning,du2026s3cot} on data with filtered \citep{li2026making,xia2025tokenskip} or rewritten tokens \citep{wu2025concise}, it remains unclear how these structurally achieve compression. For instance, CoT compression structures and their impacts on the data volume and training epochs required for SFT remains largely unexplored.

In parallel, the distinct roles of SFT and RL in LLMs \citep{chu2025sft,matsutani2026rl} have received increasing attention. A pessimistic view holds that RL merely sharpens the distribution without discovering novel solutions \citep{yue2025does}, whereas an optimistic view suggests that RL generalizes beyond SFT through unseen composition—combining skills $f_i(x)$ and $f_j(x)$ into $f_i(f_j(x))$ \citep{anderson1982acquisition,arora2023a,yuan2026from,park2025how,cheng2026from}. This compositionality is crucial for understanding RL generalization \citep{chu2025sft,shenfeld2025rlsrazor}. However, to achieve this, it remains unclear whether models can successfully decompose and reconstruct entangled skill chunks observed in imitation data.

An important question to ask is then \textit{(i) how CoT compression impacts the learning dynamics of SFT}, and \textit{(ii) whether SFT and RL can decompose compressed steps}.

To bridge this gap, we define composition in reasoning data as the bundling of multiple atomic operations into a single step, without explicitly decomposing them or emitting intermediate results. 
We first introduce a taxonomy of CoT. Specifically, we categorize CoT into fully decomposed, Explicit CoT (detailing every operation and value) and Compressed CoT (aggregating them). The latter is further subdivided into Composed CoT, which explicitly lists and executes the combined operations at once, and Implicit CoT, which yields only the final step of the chunk (\Cref{fig:main} (a)).
Based on this taxonomy, we follow \citet{ye2025PoLM21,ye2025PoLM22,zhou2025gsminfty,zhang2025on} to construct an arithmetic synthetic task with controllable difficulty, compression granularity, and CoT types to investigate different CoT data properties.

We employ Qwen2.5 \citep{yang2025qwen25} and Llama-3 \citep{grattafiori2024llama3} models (from 0.5B to 14B parameters), and analyze the effect of data size, and compression granularity, and CoT types in SFT data on out-of-distribution generalization, measured on a test set of longer compositional tasks (\Cref{fig:id_ood}), and its downstream impact on RLVR.

We found that SFT on coarsely Compressed CoT requires supervision on a larger number of data points. In particular, Compressed CoT benefits more from data scaling than Explicit CoT. Within Compressed CoT, Composed CoT benefits from data repetition, whereas Implicit CoT suffers from OOD performance degradation. We further observed that SFT cannot decompose compressed steps, while RLVR can do so via exploration, enabling generalization to longer compositional tasks that require decomposition. Finally, by analyzing CoT order, we found that unidirectional CoT generalizes, whereas hierarchically chunked CoT fails to generalize to longer tasks.

Within this controlled synthetic setting, we hope these findings will help illuminate the trade-off between reasoning length and performance and provide useful principles for data design under resource constraints.
Furthermore, we expect this work to elucidate the critical role of RL in LLM post-training, particularly how skill composition and decomposition drive generalization, ultimately helping post-training discover new solutions through skill composition.

The paper is organized as follows. In \Cref{sec:problem-setup}, we conceptually categorize CoT compression and introduce a controlled empirical task. In \Cref{subsec:data-model-granularity}, we study how compression granularity affects required SFT steps, while in \Cref{subsec:data-diversity}, we analyze data scaling and repetition across different CoT compression. In \Cref{subsec:decomposition}, we discuss reasoning chain decomposition alongside the comparative roles of SFT and RLVR. In \Cref{subsec:cot-order}, we show how CoT ordering in SFT data affects generalization. We discuss connections to prior work in \Cref{sec:discussion}. Detailed related work appears in \Cref{app:related_work}.
The code is available at \href{https://github.com/kohseim/cot_compression}{kohseim/cot\_compression}.

\section{Problem Setup}\label{sec:problem-setup}
\begin{figure*}[t]
  \centering
  \includegraphics[width=\linewidth]{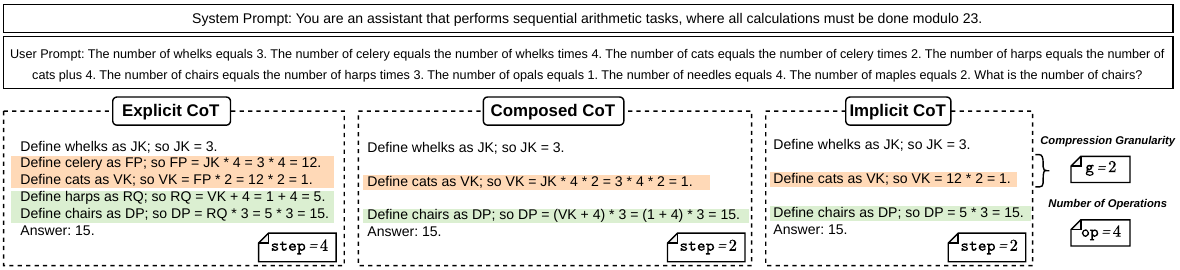}
  \caption{\textbf{Synthetic Dataset for Compositional Tasks.} Each question consists of natural language descriptions of inter-parameter relations, including addition, subtraction, and multiplication, with initial parameter values. The task requires sequentially applying the specified operations modulo 23 to infer the value of a target parameter. We use a CoT format of the form: ``Define [parameter] as [variable]; so [variable] [operation] = [value].'' This example requires 4 operations ($\texttt{op}=4$) to solve. For compressed CoT (Composed CoT and Implicit CoT), the compression granularity is set to 2 ($\texttt{g}=2$), so the problem is solved in 2 steps ($\texttt{step}=2$).}
  \label{fig:data_main}
\end{figure*}
In this section, we introduce (a) the taxonomy of CoT for compositional tasks, and (b) the synthetic tasks for experiments.

\subsection{Taxonomy of CoT}
Compositional, long sequential problems can be expressed as a chain of atomic operations. We can write this chain by nesting a transformation $f_i(\cdot)$ i.e., $f_i(f_{i-1}(\ldots (f_1(s_0)))$. For example, in math tasks, this transformation $f_i$ corresponds to mathematical operation that maps from values to calculated results. In knowledge tasks, such as multi-hop factual reasoning \citep{ho2020constructing,trivedi2022musique,press2023measuring}, $f_i$ corresponds to relation between two entities (e.g., "'s capital is" maps from "Japan" to "Tokyo"). In graph tasks \citep{wang2023can,sun2024thinkongraph}, $f_i$ corresponds to directed edge that transits from one node (vertex) to another, and in discrete state-action environments that can be described as Markov Decision Process (MDPs), such as games (e.g., spatial navigation tasks \citep{nolte2024transformers,dao2025alphamaze,li2026do} and ARC-AGI \citep{francois2019on}) and robotics tasks, $f_i$ corresponds to an action (e.g., "Up", "Down") that moves the agent (LLM) from one state to another. Hereafter, we refer to $f_i$ as an \emph{operation} and $s_i$ as a \emph{value}.

We then formalize the compression of CoT. Prior works proposed methods to make compressed reasoning data by filtering tokens \citep{li2026making,xia2025tokenskip}, rewriting reasoning expressions \citep{wu2025concise}, and self-distillation \citep{huang2026reasoning,du2026s3cot}, but these methods remain heuristic and lack general validity across models and domains.
Accordingly, we categorize CoT into \emph{Explicit CoT}, \emph{Composed CoT}, and \emph{Implicit CoT}.

\paragraph{Explicit CoT}
Explicit CoT processes each operation step by step without aggregating or skipping them: $s_1 = f_1(s_0)$, $s_2 = f_2(s_1)$, $\ldots$, where $s_i$ is an (intermediate) value.

\paragraph{Composed CoT}
In Composed CoT, multiple operations are composed within a single reasoning step. Thus, the model proceeds as $s_2 = f_2(f_1(s_0)), s_4 = f_4(f_3(s_2)), \ldots$ Composed CoT omits intermediate values while explicitly specifying all applied operations.
In this case, we call the \emph{compression granularity} of CoT is 2 ($\texttt{g}=2$), two functions are composed into one step, and we control this granularity hereafter.
Explicit CoT corresponds to $\texttt{g}=1$.

\paragraph{Implicit CoT}
Implicit CoT skips operations, so the model only outputs $s_4 = f_4(s_3)$, and skips $s_1 = f_1(s_0)$, $s_2 = f_2(s_1)$, and $s_3 = f_3(s_2)$, where compressed granularity is 4 ($\texttt{g}=4$). While compressed CoT outputs all the operations they apply and hide the values, Implicit CoT hides and internally processes both the operations and the values. Implicit CoT can substantially reduce response length, which has motivated methods for internalizing CoT into continuous states \citep{deng2023implicit,deng2024explicit,hao2025training,shen2025codi,wei2026simcot}, but their learning limitations have been noted \citep{li2026chain}.

\begin{figure*}[t]
  \centering
  \includegraphics[width=\linewidth]{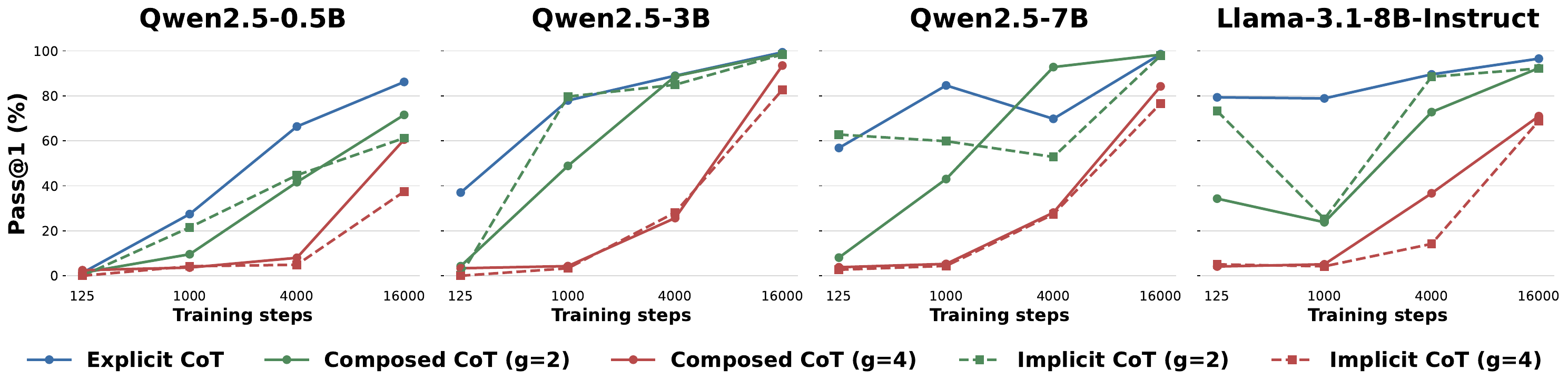}
  \caption{\textbf{Compression Granularity and Training Steps.} The bar chart reports the average performance of Qwen2.5-0.5B, 3B, 7B, and Llama-3.1-8B-Instruct at steps 125, 1000, 4000, and 16000 after SFT with Explicit CoT, Composed CoT, and Implicit CoT with $\texttt{g}=2,4$. Models are trained on tasks with $\texttt{op}=8,16,24$. Evaluation results are averaged over $\texttt{op}=32,40,48,\ldots,96,104$ tasks.}
  \label{fig:granularity_data}
\end{figure*}

\subsection{Synthetic Dataset for Compositional Tasks}
\begin{figure}[H]
  \centering
  \includegraphics[width=\linewidth]{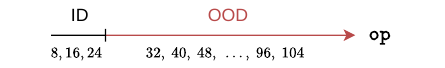}
  \caption{\textbf{Train–Test Split by $\texttt{op}$.} Training is performed on tasks with short $\texttt{op}$ sequences, while tasks with longer $\texttt{op}$ sequences are used for OOD evaluation.}
  \label{fig:id_ood}
\end{figure}
For these compositional reasoning tasks, we construct a testbed that lets us control data size, difficulty, and compression granularity. We employ synthetic arithmetic dataset illustrated in \Cref{fig:data_main}. Parameter dependencies are given in the contexts. Each dependency involves a single operation from $\{+, -, \times\}$, and we define the number of such operations as the difficulty, denoted by \texttt{op}. We define out-of-distribution (OOD) tasks as those with larger \texttt{op} than in the training set following \citet{zhang2025on} (\Cref{fig:id_ood}). This arithmetic task is formulated as a "chaining" of many atomic operations with only local dependencies.
This dataset is a variant of those in \citet{ye2025PoLM21,ye2025PoLM22,zhou2025gsminfty,zhang2025on}. Our task corresponds to the special case where the in-degree and out-degree of their computational graph are both 1. In addition, to prevent numerical explosion as \texttt{op} grows, we restrict the operations to modulo 23 following \citep{ye2025PoLM21,ye2025PoLM22}. To more closely approximate a real-world setting, we follow \citet{zhou2025gsminfty} and inject noise parameters into the context. See \Cref{app:synthetic_dataset} for detailed data generation process.

\section{Experiments}\label{sec:experiments}
The interplays between data points and compressed granularity are addressed in \Cref{subsec:data-model-granularity}. Data diversity considered in \Cref{subsec:data-diversity}, decomposition of Compressed CoT is investigated in \Cref{subsec:decomposition}, and difference between CoT orders are addressed in \Cref{subsec:cot-order}. For details on SFT and RLVR training, please refer to \Cref{app:training}.

\paragraph{Notations.}
$\texttt{op}$ denotes the number of operations required to solve a compositional task.
CoT is categorized into \emph{Explicit CoT} and \emph{Compressed CoT}, where the latter includes \emph{Composed CoT} and \emph{Implicit CoT}.
For Compressed CoT, $\texttt{g}$ denotes the number of operations grouped into a single step, and $\texttt{step}$ denotes the number of steps required to solve the task.
Thus, it holds that $\texttt{op} = \texttt{g} \times \texttt{step},$ where $\texttt{g}=1$ for \emph{Explicit CoT}.

\begin{figure*}[!t]
  \centering
  \includegraphics[width=\linewidth]{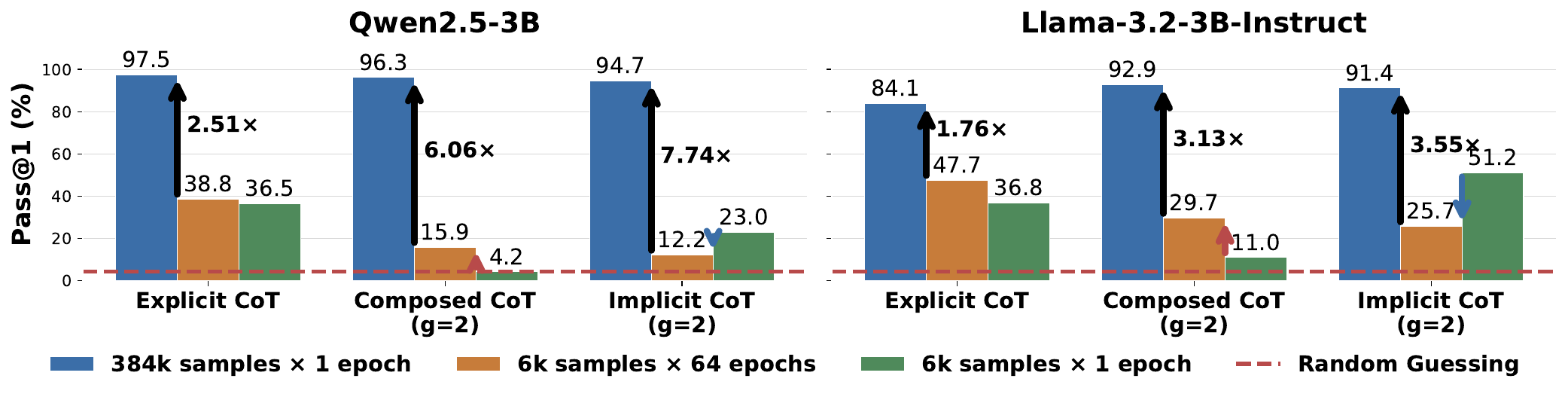}
  \caption{\textbf{Data Scaling vs Data Repetition.} The bar chart reports average performance of Qwen2.5-3B and Llama-3.2-3B-Instruct after SFT with Composed CoT and Implicit CoT with $\texttt{g}=2$. Models are trained with 384k samples for 1 epoch, 6k samples for 64 epochs, and 6k samples for 1 epoch. Evaluation results are averaged over $\texttt{op}=32,40,48,\ldots,96,104$ tasks. Since the computation is performed modulo 23, the chance level ($\frac{1}{23}$), is indicated by the red dashed line.}
  \label{fig:data-diversity}
\end{figure*}
\subsection{Compression Granularity and Training Steps}\label{subsec:data-model-granularity}
We first analyze how compressed reasoning data affect SFT, particularly on how the compression granularity influences the number of training steps required for performance improvement.

We perform SFT on Qwen2.5-0.5B, 1.5B, 3B, 7B, 14B, and  Llama-3.2-1B, 3B-Instruct and Llama-3.1-8B-Instruct using Explicit CoT, Composed CoT with $\texttt{g}=2,4$, and Implicit CoT on $\texttt{op}=8,16,24$ tasks. Training is conducted for one epoch, corresponding to 16,000 steps, on 768k samples. Since SFT is performed only on $\texttt{op}$ that are multiples of 8, and multiples of the compression granularities considered here (2 and 4), the model can reach the final answer without fractional operations.
We evaluate each model on $\texttt{op}=32,40,48,\ldots,104$ tasks, where each task has 336 samples. Throughout the rest of this paper, evaluation is performed with greedy decoding. The maximum number of tokens is set by the $\texttt{op}$ range: 4096, 8192, 12288, and 16384 for $\texttt{op}=25$-44, 45-64, 65-84, and 85-104, respectively, ensuring sufficient generation length.

\Cref{fig:granularity_data} shows performance (Pass@1) averaged over $\texttt{op}=32,40,48,\ldots,96,104$ at each training step for each model and SFT setting. Across all model sizes, larger $\texttt{g}$ in Compressed CoT, including both Composed CoT and Implicit CoT, requires more training steps to reach the same level of performance. One can interpret this as evidence that, to train models to use CoT with higher compression granularity (coarser reasoning traces), one should prepare and perform SFT on more data points. See \Cref{app:data_granularity} for detailed results.

\paragraph{Math Experiments}
To bridge it to real-world tasks, we conduct experiments on math. We rewrite the reasoning traces in train set of Hendrycks-MATH \citep{hendrycks2021measuring} (level 3-5) into Composed CoT and Implicit CoT using Qwen3.5-Flash \citep{qwen3.5} with the prompts describing each CoT, perform SFT on Qwen2.5-7B for 2 epochs, and measured pass@1 (\%) on GSM8K \citep{cobbe2021training}, MATH500 \citep{hendrycks2021measuring}, and AMC23 with greedy decoding. See \Cref{app:data_granularity} for the detailed settings.

\begin{figure}[h]
  \centering
  \includegraphics[width=\linewidth]{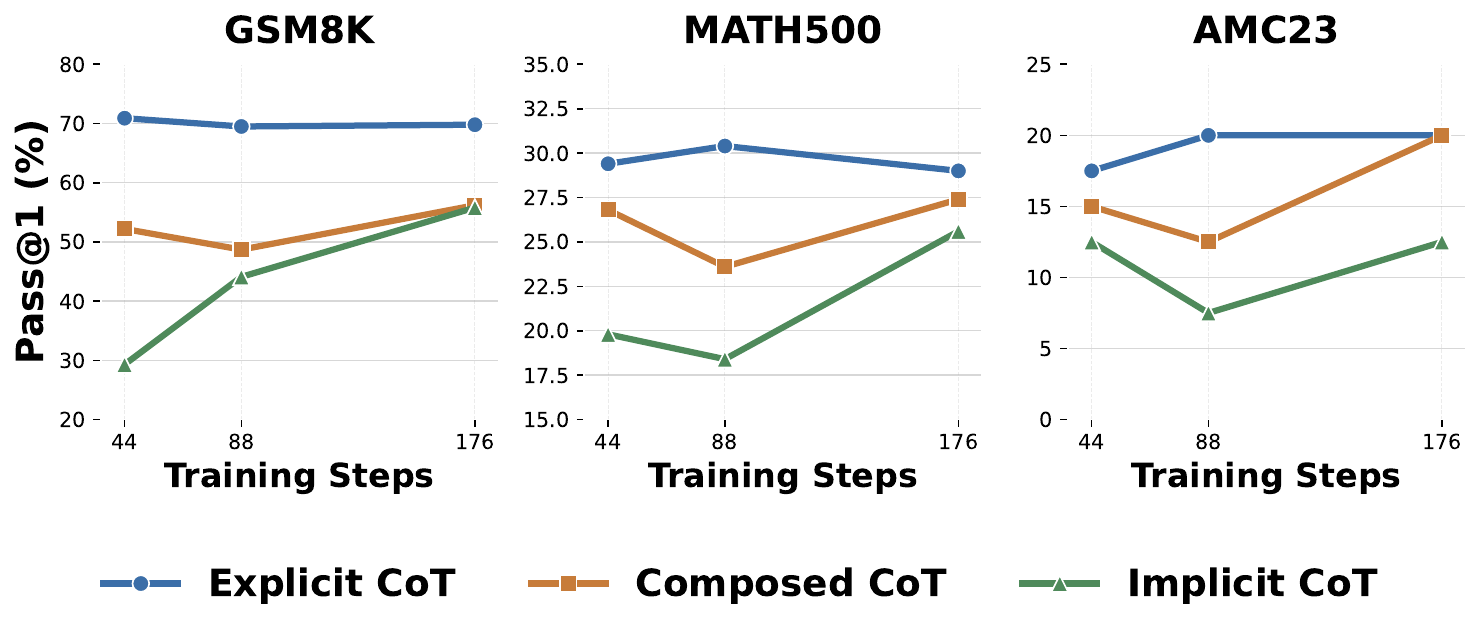}
  \caption{\textbf{Compressed Data in Math.} The bar chart reports the average performance of Qwen2.5-7B after SFT on Explicit CoT (the original reasoning traces) and on Composed CoT and Implicit CoT rewritten by Qwen3.5-Flash.}
  \label{fig:math}
\end{figure}

\Cref{fig:math} show that Composed CoT and Implicit CoT requires more training data than Explicit CoT in a math domain.

\begin{takeaway}
\paragraph{Takeaway 1.} Coarser-grained Compressed CoT requires more data.
\end{takeaway}

\subsection{Data Scaling vs Data Repetition}\label{subsec:data-diversity}
\begin{figure*}[!t]
  \centering
  \begin{subfigure}[t]{0.36\linewidth}
    \centering
    \includegraphics[width=\linewidth]{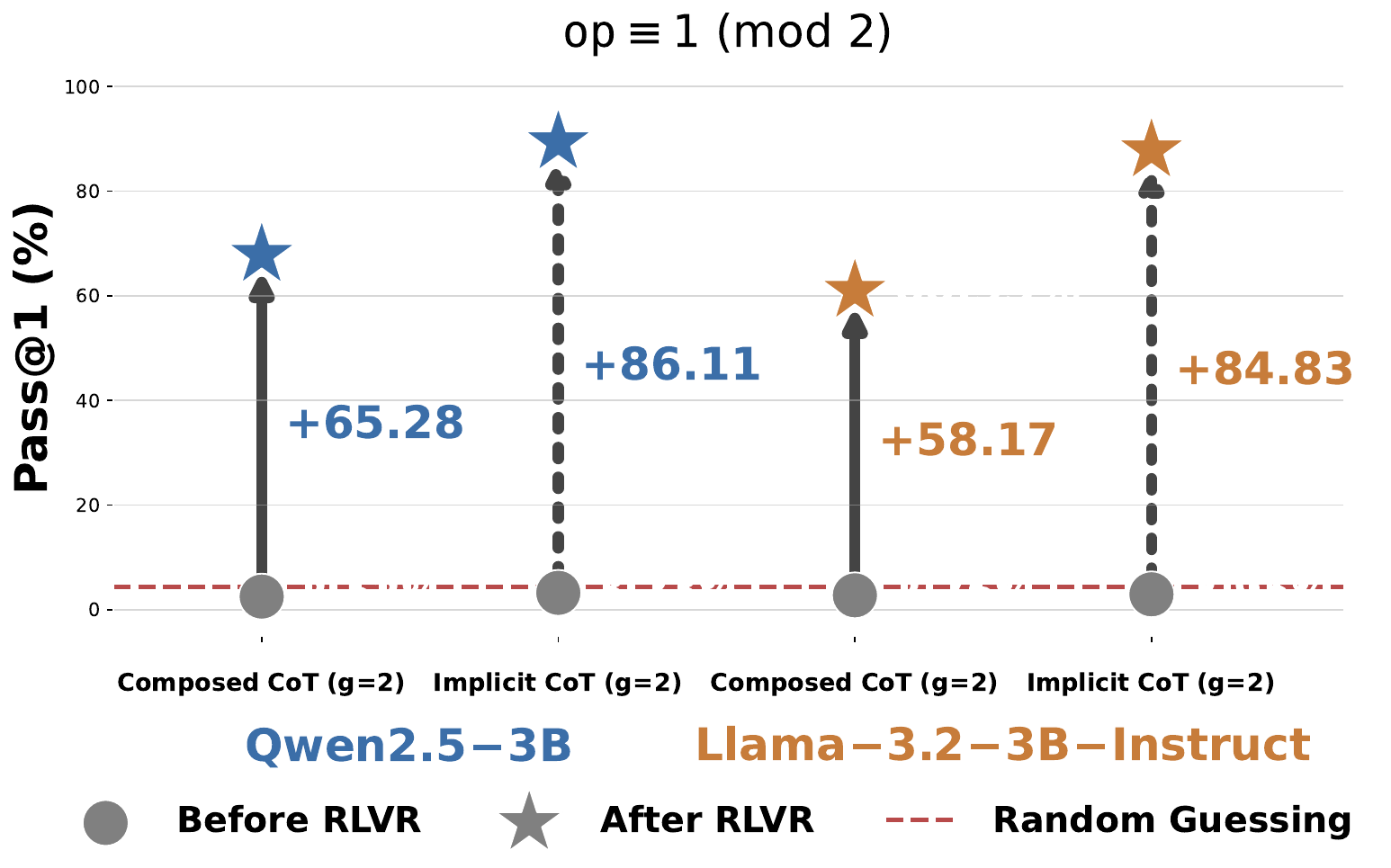}
    \caption{RLVR Evaluation Results.}
    \label{fig:rlvr_eval_main}
  \end{subfigure}
  \hspace{0.005\linewidth}
  \begin{subfigure}[t]{0.60\linewidth}
    \centering
\includegraphics[width=\linewidth]{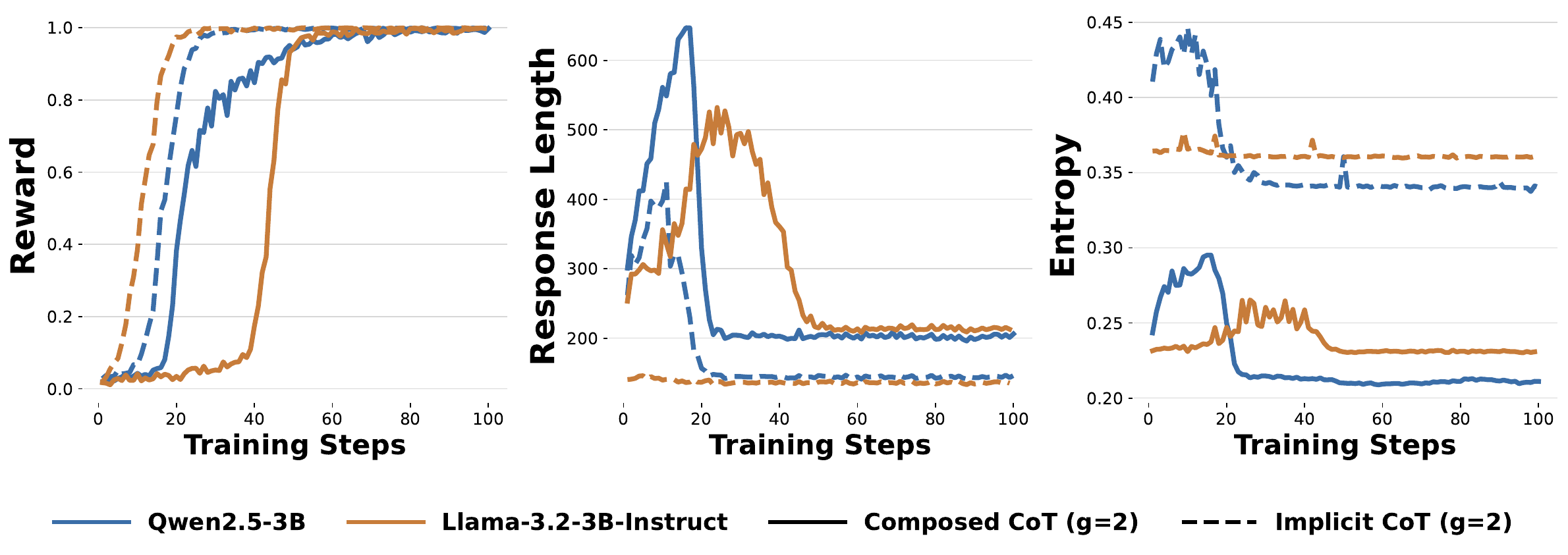}
    \caption{RLVR Training Dynamics.}
    \label{fig:rlvr_reward_curve_sub}
  \end{subfigure}
  \caption{\textbf{Decomposition of Composed Steps by RLVR.} (a) Dumbbell plot showing changes in the average evaluation results over $\texttt{op}=25,27,29,\ldots,101,103$ before and after RLVR on $\texttt{op}=9,11,13,15$ tasks, using checkpoints obtained by SFT on Qwen2.5-3B and Llama-3.2-3B-Instruct with Composed CoT and Implicit CoT with $\texttt{g}=2$. (b) Training dynamics of the mean reward, mean rollout response length, and mean token entropy at each steps. Since the computation is performed modulo 23, the chance level ($\frac{1}{23}$), is indicated by the red dashed line.}
  \label{fig:rlvr_reward_curve}
\end{figure*}

Having analyzed how compression granularity affect the number of SFT training steps required, we next investigate whether data repetition or scaling is more effective for Compressed CoT data. This analysis is motivated by prior work showing that, in long CoT SFT, repeating a small amount of high-quality data can outperform scaling to larger and more diverse datasets \citep{ye2025limo,muennighoff2025s1,kopiczko2026data}. 

We perform SFT on Qwen2.5-3B and Llama-3.2-3B-Instruct under three training settings, with 384k samples for 1 epoch, 6k samples for 64 epochs, and 6k samples for 1 epoch. Note that the settings of 384k samples for one epoch and 6k samples for 64 epochs have the same computational budget. We evaluate each model on $\texttt{op}=32,40,48,\ldots,104$ tasks. Comparing 384K samples for 1 epoch with 6K samples for 64 epochs, Composed CoT and Implicit CoT show larger relative accuracy gains from diverse data scaling than Explicit CoT. This suggests that Compressed CoT requires data scaling more than Explicit CoT.

\Cref{fig:data-diversity} shows performance averaged over $\texttt{op}=32,40,48,\ldots,96,104$ for each model and SFT setting. Since answers are integers in ${0,\ldots,22}$ for our modulo 23 arithmetic task, we compare relative performance. 

\begin{takeaway}
\paragraph{Takeaway 2.} Compared with Explicit CoT, Composed CoT benefits more from diverse data than from repeated data under a fixed computational budget.
\end{takeaway}

Additionally, comparing 6K samples for 64 epochs with 6K samples for 1 epoch shows that data repetition improves Composed CoT but degrades Implicit CoT. Composed CoT explicitly represents the operation applied at each step as
$f_i(f_{i-1}(\cdot))$. In contrast, Implicit CoT depicts only the latter component, $f_i(\cdot)$. These results suggest that Implicit CoT may be more prone to memorization or overfitting to a specific length \citep{dziri2023faith,pruthi2026why}. One can benefit from SFT with Compressed CoT, which reduces response length, when diverse data are available. In contrast, when data are limited, one should apply data repetition for Composed CoT, whereas such repetition should be avoided for Implicit CoT.
See \Cref{app:data_diversity} for detailed results.

\begin{takeaway}
\paragraph{Takeaway 3.} SFT on Composed CoT improves performance through data repetition, whereas Implicit CoT exhibits performance degradation.
\end{takeaway}

\subsection{Decomposition of Reasoning Chains}\label{subsec:decomposition}
\begin{figure}[H]
  \centering
  \includegraphics[width=\linewidth]{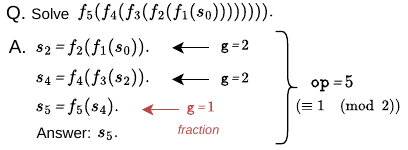}
  \caption{\textbf{Illustration of Tasks Requiring Decomposition.} For tasks with $\texttt{op}=5$ $(\equiv 1 \pmod{2})$, applying CoT with $\texttt{g}=2$ produces $\texttt{g}=1$ fractions required to solve the problem.}
  \label{fig:decompostition_illustration}
\end{figure}
\begin{figure}[t]
  \centering
  \includegraphics[width=\linewidth]{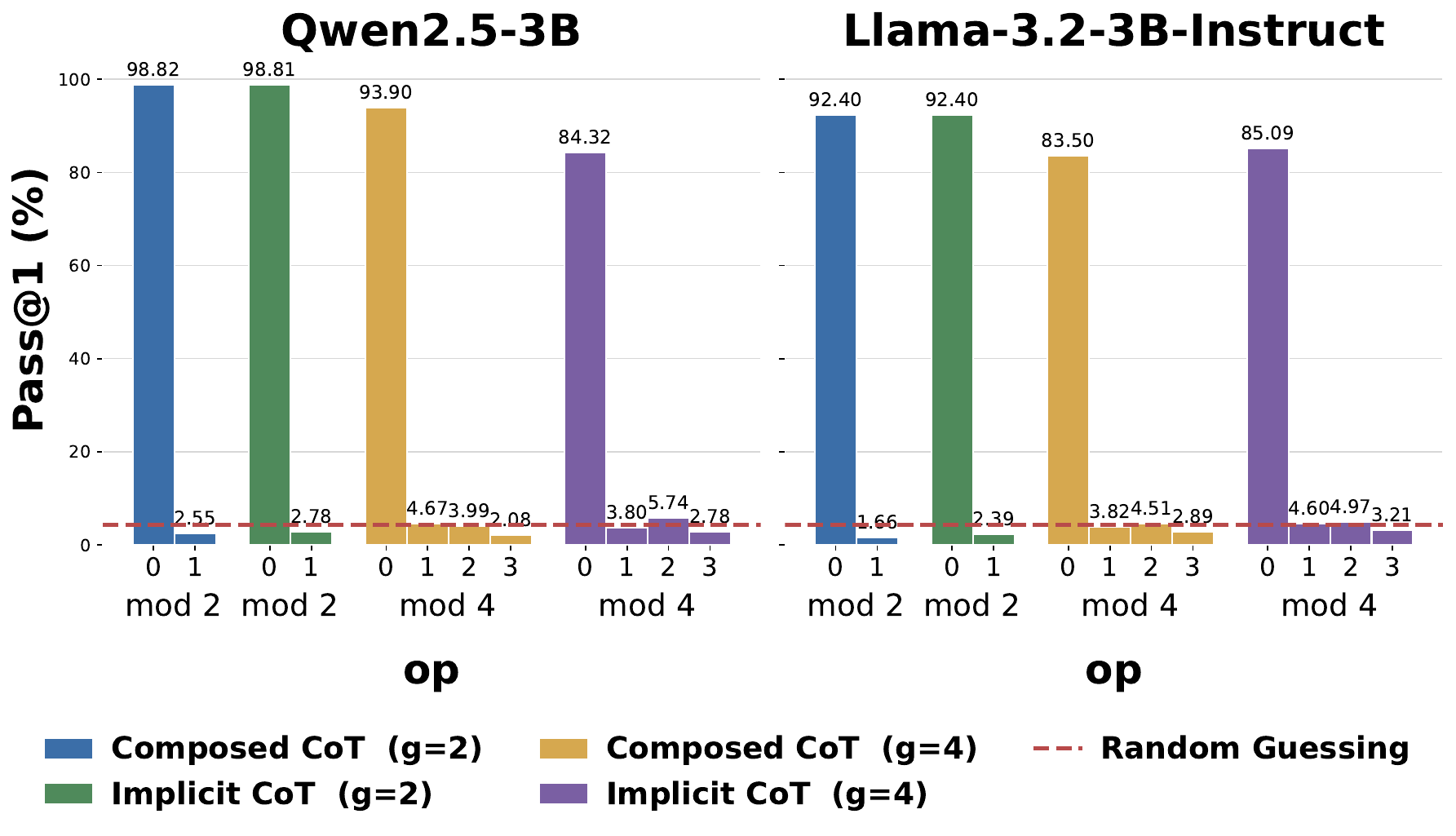}
  \caption{\textbf{Decomposition of Composed Steps by SFT.} The bar chart reports the average performance of Qwen2.5-3B and Llama-3.2-3B-Instruct after SFT on 384k samples using Composed CoT and Implicit CoT with $\texttt{g}=2,4$, evaluated on $\texttt{op}=25,26,27,\ldots,103,104$. Results are averaged over $\texttt{op}$ values grouped by residue classes modulo 2 ($0,1$), and modulo 4 ($0,1,2,3$). Since the computation is performed modulo 23, the chance level ($\frac{1}{23}$), is indicated by the red dashed line.}
  \label{fig:sft_decompostition}
\end{figure}

We investigate (i) whether SFT can decompose reasoning steps below the compression granularity, and (ii) whether such decomposition can be induced, particularly through subsequent RLVR. Prior work has shown that, for atomic operations (skills) already acquired by LLMs, SFT fails to handle unseen compositions, whereas RL generalizes to them \citep{yuan2026from,park2025how,cheng2026from}. However, enabling such composition requires decomposing aggregated operations observed in imitation data and recombining the resulting components \citep{lake2018generalization,kim2020cogs,keysers2020measuring,hupkes2021compositionality}.
\begin{figure*}[!t]
  \centering
  \includegraphics[width=\linewidth]{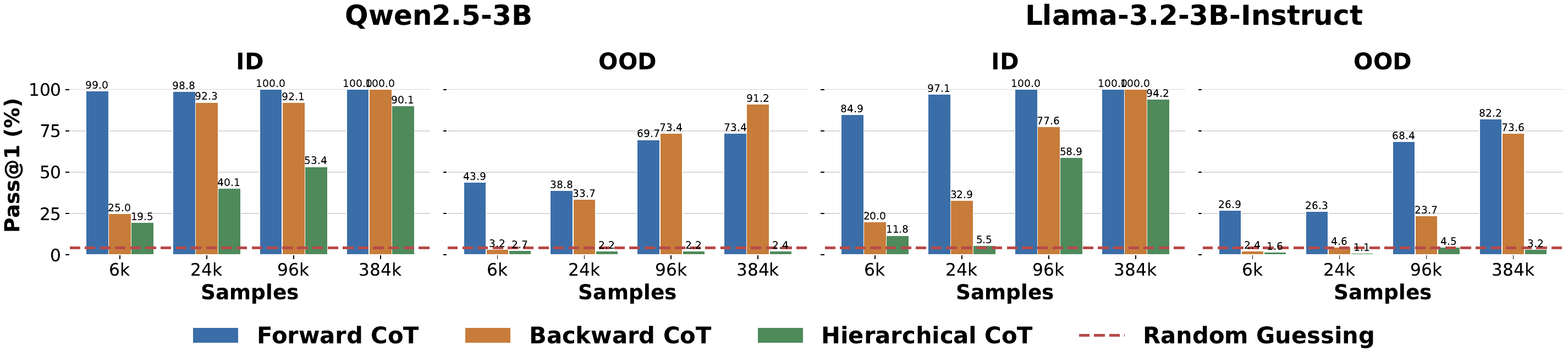}
  \caption{\textbf{Effect of CoT Order.} The bar chart reports the average performance of Qwen2.5-3B and Llama-3.2-3B-Instruct after SFT on 384k samples using Forward CoT, Backward CoT, and Hierarchical CoT with $\texttt{op}=8,16$, evaluated on $\texttt{op}=8,16$ (ID) and $\texttt{op}=32,64,128$ (OOD). Since the computation is performed modulo 23, the chance level ($\frac{1}{23}$), is indicated by the red dashed line.}
  \label{fig:cot_order}
\end{figure*}
To this end, we evaluate it on problems that require decomposition, as their computation cannot be represented exactly at the given compression granularity (\Cref{fig:decompostition_illustration}). Specifically, for Qwen2.5-3B and Llama-3.2-3B-Instruct, we perform SFT on CoT traces with $\texttt{g}=2$, so that the smallest observed unit is $f_i(f_{i-1}(\cdot))$, while $f_i$ itself is never observed in isolation. We test whether the models can solve problems that require decomposition into individual $f_i$ steps, and whether subsequent on-policy RLVR enables this ability without access to offline decomposed CoT traces.

We train the models for one epoch on 384k samples with $\texttt{op}=8,16,24$ ($\equiv 0 \pmod{2}$), and evaluate them on $\texttt{op}=25,27,29,\ldots,101,103$ ($\equiv 1 \pmod{2}$). We then perform GRPO \citep{shao2024deepseekmath} on $\texttt{op}=9,11,13,15$ $(\equiv 1 \pmod{2})$ and evaluate on $\texttt{op}=25,27,29,\ldots,101,103$ tasks.

\Cref{fig:sft_decompostition} reports evaluation results after SFT on even and odd $\texttt{op}$. The results show that, when trained on compressed CoT traces with $\texttt{g}=2$, both Composed CoT and Implicit CoT solve OOD tasks with even $\texttt{op}$ but fail on tasks with odd $\texttt{op}$.

\begin{takeaway}
\paragraph{Takeaway 4.} SFT cannot decompose reasoning steps below the compression granularity of the training data.
\end{takeaway}

However, applying RLVR to problems with odd $\texttt{op}$ enables the models to solve tasks that were almost unsolved after SFT, where performance remained near chance level. As shown in \Cref{fig:rlvr_reward_curve}, the models acquire the ability to solve problems that require decomposition. This shows that decomposition can be achieved using only outcome correctness, without preparing CoT traces as in SFT. It also suggests that, with subsequent RLVR, hidden steps learned through compressed SFT data can be used in new compositions.
The response length dynamics further support this interpretation. Response length increases sharply when the reward first starts to rise, and then decreases as training converges. This phenomenon is not observed when RLVR is applied to even $\texttt{op}$ tasks (\Cref{fig:eval_rlvr_odd_even}). This may indicate that the model first decomposes the $\texttt{g}=2$ compressed reasoning into explicit CoT steps, and then converges to a more efficient strategy that uses $\texttt{g}=2$ compressed CoT for the main computation and a single step operation ($\texttt{g}=1$) for the fraction. See \Cref{app:rlvr} for examples of reasoning outputs.

Moreover, Implicit CoT requires fewer optimization steps than Composed CoT before the reward starts to increase. Composed CoT exhibits higher average token entropy along trajectories, indicating that it is more strongly constrained by the SFT format of emitting the observed $f_i(f_{i-1}(\cdot))$ chunks under $\texttt{g}=2$. See \Cref{app:rlvr} for the RLVR baselines and detailed results.

\begin{takeaway}
\paragraph{Takeaway 5.} RLVR enables decomposition thanks to its on-policy exploration. Implicit CoT can be decomposed more efficiently than Composed CoT with RLVR.
\end{takeaway}

\subsection{Effect of CoT Order on Generalization}\label{subsec:cot-order}

\begin{figure}[H]
  \centering
  \includegraphics[width=\linewidth]{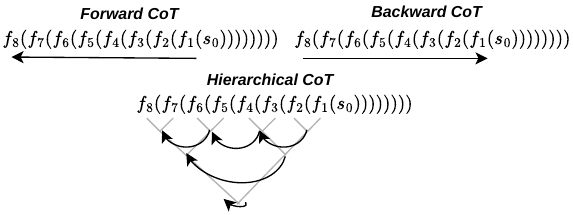}
  \caption{\textbf{Illustration of Different CoT Orders.} Forward CoT, Backward CoT, and Hierarchical CoT for tasks with $\texttt{op}=8$.}
  \label{fig:cot_order_illustration}
\end{figure}

We have so far examined compressed CoT mainly for sequential problems such as $f_8(f_7(f_6(f_5(f_4(f_3(f_2(f_1(s_0))))))))$, where $\texttt{op}=8$ and CoT follows the forward order $f_1 \to f_2 \to f_3 \to f_4 \to f_5 \to f_6 \to f_7 \to f_8$ (we call this \emph{forward CoT}.)
Other reasoning orders are also possible. In \emph{backward CoT}, the model reasons backward from the desired answer, following $f_8 \to f_7 \to f_6 \to f_5 \to f_4 \to f_3 \to f_2 \to f_1$. In \emph{Hierarchical CoT}, the problem is split into chunks (sets of adjunct functions), intermediate results are stored as variables, and these results are composed hierarchically. Intuitively, this corresponds to solving a mathematical problem by evaluating useful subexpressions and then combining them to obtain the final answer. Illustrations of different CoT orders are highlighted in \Cref{fig:cot_order_illustration}.

To quantify how these CoT orders affect SFT, we train for one epoch on 6k, 24k, 96k, and 384k samples with $\texttt{op}=8,16$. We evaluate ID performance at $\texttt{op}=8,16$ and OOD performance at $\texttt{op}=32,64,128$. For Hierarchical CoT, we set $\texttt{op}$ to powers of two to ensure a binary-tree structure.

\Cref{fig:cot_order} shows the average ID and OOD evaluation results for each CoT order. For both models, Hierarchical CoT fails to generalize to OOD (longer compositional tasks), whereas Forward CoT and Backward CoT do generalize. However, Backward CoT requires more data. This suggests that sequential reasoning in a unidirectional order is important for long CoT in Transformer-based LLMs, and that the larger data requirement of Backward CoT during post-training may be related to the data distribution used in pre-/mid-training. Intuitively, non-unidirectional CoT, such as Hierarchical CoT, may require more advanced reasoning and greater working memory, as it must retain intermediate values from multiple chunks and use them in future operations.
When constructing a CoT trace for sequential tasks, one should preferably adopt a unidirectional design in which a single value is used in the subsequent operation. See \Cref{app:cot_order} for detailed results.

\begin{takeaway}
\paragraph{Takeaway 6.} Forward CoT and Backward CoT are more robust on sequential compositional tasks, compared to Hierarchical CoT. Backward CoT requires more data than Forward CoT.
\end{takeaway}

\section{Discussion and Related Works}\label{sec:discussion}

In this work, we study when and how compressed reasoning data affect SFT and subsequent RLVR. We first find that increasing compression granularity requires more training steps and more data to achieve performance gains. Among the CoT formats, Compressed CoT benefits more from data scaling, and within Compressed CoT, Composed CoT benefits from data repetition, whereas Implicit CoT is harmed by it. We further show that, even when models are exposed to compressed reasoning data during SFT, subsequent RLVR exploration can decompose reasoning into atomic operations. Finally, we analyze the ordering of CoT supervision and show that Hierarchical CoT generalizes poorly OOD compared with sequential Forward and Backward CoT, while Backward CoT requires more data than Forward CoT. Below, we discuss connections to related literature, practical implications, and potential directions for future work.

\paragraph{Accuracy and Length Pareto Frontier.}

As LLM agents are deployed across a wide range of use cases, post-training on compressed reasoning data becomes increasingly important for inference efficiency and token cost \citep{bai2026how}. Our results demonstrate that compressed CoT requires more data points, and that an SFT-then-RL pipeline is effective \citep{matsutani2026rl,limozin2026sftthenrl}.

\paragraph{Synthetic Reasoning Data.}

Our results also provide practical guidelines for designing compressed reasoning data under limited resource constraints. For example, when training with short CoT data via SFT but data diversity is limited, it is preferable to train for multiple epochs on Composed CoT, which explicitly includes all operations at a single steps, rather than on Implicit CoT. More broadly, promising directions include designing synthetic data generation pipelines that produce appropriate CoTs from problems \citep{wu2025concise}, as well as rewriting existing reasoning traces \citep{su2025nemotron,fujii2026rewriting} into effective compressed formats.

\paragraph{Composition in LLMs.}
We find that SFT on compressed reasoning data cannot decompose reasoning below the granularity observed in supervision, whereas subsequent RLVR can recover atomic operations and generalize to longer compositional tasks. Unlike SFT, RLVR does not require explicitly decomposed CoT traces. 
For example, supervision on only $f_2(f_1(\cdot))$ and $f_3(f_2(\cdot))$ may suffice to recover the underlying components $f_1$, $f_2$, and $f_3$, and potentially recombine them to form $f_1(f_3(\cdot))$.
In our synthetic task, we consider only three operations, $\{+, -, \times\}$. In real-world domains such as mathematics, code, science, and logical reasoning, however, the space of effective operations is much larger, and often semi-infinite. Because operation combinations grow combinatorially, manually decomposing them during data construction is often impractical, making it important for an RL algorithm to discover useful decompositions from outcome reward supervision alone.
This observation echoes prior findings that SFT struggles with unseen compositions, whereas RL can generalize to OOD compositions \citep{yuan2026from,park2025how,cheng2026from}. Against recent pessimistic views on RL for LLMs \citep{yue2025does}, our results suggest that RL can break apart coarse reasoning chunks learned from compressed SFT data and recombine them into further compositions, supporting the view that RL can discover new solutions not present in the base model. 
We believe that composition and decomposition through RL are a promising direction toward LLMs that learn from experience \citep{silver2025welcome} with less dependence on data coverage \citep{chen2026the,zhang2025on}.

\paragraph{Internal Circuits for Compositional Reasoning.}
Prior work has studied internal circuits in multi-hop reasoning of LLMs \citep{li2024understanding,yang2024large,biran2024hopping,yu2025back,tang2025explainable,hong2026a,yao2026compositional}. It would be interesting to study compositional reasoning in models trained with Explicit CoT versus Compressed CoT, via mechanistic interpretability analyses of internal computation.

\section*{Limitations}
This study has several limitations. First, our experiments rely on synthetic arithmetic tasks to control the experimental setting, and we do not perform extensive experiments on real-world datasets at large scale. Second, our formulation of CoT reasoning focuses on compositional tasks with nested operations. It does not cover tasks with more diverse structures, such as branching procedures or search. Third, we define OOD evaluation as generalization to longer compositional tasks. This notion of OOD differs from domain shift, where the input distribution changes in content or domain. Finally, our experiments are conducted with Transformer-based Qwen2.5 and Llama 3 models. We do not evaluate whether the findings extend to a broader range of architectures, such as Mamba \citep{gu2024mamba}, GatedDeltaNet \citep{yang2025gated}, hybrid models \citep{lenz2025jamba,hu2026griffin}, and Looped Transformers \citep{giannou2023looped}.

\section*{Acknowledgments}
This work was supported by the AI Symbiotic Future Society Program at The University of Tokyo.

\bibliography{custom}

\appendix

\label{sec:appendix}

\section*{AI Assistant Usage}
We used LLMs to obtain feedback on ideas and experiments, and for rephrasing written text and checking grammar. We also used LLMs for coding and debugging. We did not use LLMs to make claims made in the paper.

\section{Extended Related Work}\label{app:related_work}

\paragraph{RL vs SFT.}
\citet{yue2025does} evaluates the Pass@k metric \citep{chen2021evaluating,song2025mind,dang2025weight,wen2026reinforcement,wu2025invisible} and shows that RLVR primarily sharpens the base model distribution rather than expanding beyond its support \citep{dang2025weight,cui2025entropy,matsutani2026rl}. Several studies \citep{walder2025passk,takashiro2026advantage,parmas2026ordergrad} have explored exploration for policy-gradient methods.
\citet{wang2025octothinker,zhang2026learning,chen2026the,zhang2025on} also observed that RLVR is effective when sufficient coverage is already established during mid-training. By contrast, \citet{chu2025sft} has found that RL tends to generalizes to OOD and mitigates catastrophic forgetting, which has been interpreted as finding solutions that minimize distributional divergence \citep{shenfeld2025rlsrazor,chen2025retaining}.
For reasoning SFT, data diversity and distributional coverage are argued to drive generalization \citep{huan2025does,lin2025debunk,ren2026rethinking}, while repeated training on high-quality data can also be effective \citep{ye2025limo,muennighoff2025s1,kopiczko2026data}. Reasoning trajectories matter beyond correctness \citep{gandhi2024stream,chandra2026shape}, so does their amenability to RL \citep{zhang2026good}.

\paragraph{Long CoT and Reasoning Compression.}
LLMs can enhance their reasoning ability by producing Long CoT \citep{yeo2025demystifying}. Prior work has analyzed several aspects of Long CoT reasoning, including planning \citep{gregor2024the,nagarajan2025roll}, search \citep{gandhi2024stream,gandhi2025cognitive,saparov2025transformers}, self-correction \citep{huang2024large,tyen2024llms}, and granularity \citep{wang2026beyond}. However, as task complexity increases, Long CoT can suffer from snowballing errors \citep{dziri2023faith,shojaee2026the} and overthinking \citep{sui2025stop}. To address these issues, recent studies have explored methods for compressing reasoning traces, such as filtering tokens \citep{li2026making,xia2025tokenskip}, prompting teacher models \citep{wu2025concise}, to generate compressed CoT traces for SFT, and for self-distillation \citep{huang2026reasoning,du2026s3cot} or on-policy distillation \citep{sang2026crisp}.
Methods have been proposed to internalize CoT traces into continuous states, which are referred to as Latent CoT \citep{deng2023implicit,deng2024explicit,hao2025training,shen2025codi,wei2026simcot}.
RL approaches have also been incorporated response length penalties into the reward \citep{hou2026thinkprune,huang2026mitigating,chen2026theoverthinker,liu2026learn,liu2025dler}. 

\paragraph{Composition in LLMs.}

\citet{arora2023a} formalized skill compositons in LLMs. Building on this, previous works studied data selection \citep{chen2023skillit}, in-context prompting \citep{chen2024skills}, benchmark \citep{yu2024skillmix}, SFT \citep{zhao2024can}, especially by conditioning behaviors \citep{didolkar2025metacognitive} and targeting skills \citep{he2025skill}, and self-distillation \citep{sprague2026skillfactory}. \citet{yao2026compositional} analyzed compositional generalization under distribution shift and \citet{lippl2025algorithic} identified compositional geometry of algorithmic primitives. Recently, \citet{yuan2026from} and \citet{cheng2026from} investigated the effect of RL on compositional abilities, using synthetic tasks with nested functions and synthetic biographical data, respectively, and \citet{park2025how} employed countdown tasks \citep{gandhi2024stream} to explain structure-dependent hierarchy of learnability.\\

\section{Synthetic Dataset}\label{app:synthetic_dataset}

\paragraph{Data Generation Framework.}

To control data size, difficulty, and compression granularity in compositional reasoning tasks, we construct a GSM8K-like \citep{cobbe2021training} arithmetic task following \citet{liu2023tinygsm, li2024gsmplus,zhou2025gsminfty,mirzadeh2025gsmsymbolic,zhang2025on}. In data generation, we specify the number of operations required to solve the task, denoted by \texttt{op}, and choose the CoT type of the reasoning trace from Explicit CoT, Composed CoT, and Implicit CoT. For Composed CoT and Implicit CoT, which we collectively refer to as Compressed CoT, we additionally specify the compression granularity \texttt{g}, namely the number of operations processed in a single reasoning step. We denote the resulting number of CoT steps by \texttt{step}.

We summarize below the primary sources of randomness in the data generation process.

\begin{itemize}
    \item Initial value: The value of the first parameter is randomly sampled from $\{1,2,3,4\}$.
    \item Operation: The dependency between two adjacent parameters is randomly sampled from $\{+, -, \times\}$.
    \item Parameter name: Parameters are randomly allocated from 200 populations without duplication within the same dataset.
    \item Variable name: Variables are assigned random two-letter labels (e.g., DL, ML, and AI). There are $26^2=676$ possible combinations.
    \item Number of distractor parameters: The number of distractor parameters is randomly sampled from \(\{1, 2, \ldots, \texttt{op}-1\}\).
    \item Distractor parameter value: The value of each distractor parameter is randomly sampled from $\{1,2,3,4\}$.
    \item Sentence permutation: The problem statement jointly permutes all \(\texttt{op} + 1\) parameters and distractor parameters.
\end{itemize}

In our experiments, data are mixed so that the number of examples is balanced across the different values of \texttt{op} used for training and evaluation. We also preprocess the data to ensure that there is no duplication between the training and evaluation sets. Moreover, owing to the randomness of the generation process, we can generate a combinatorial space of examples that is far larger than the scale of the SFT data used in this study (up to 768k examples in a single run).

To prevent numerical values from growing explosively when using long operation sequences with a large number of operations \texttt{op} over $\{+, -, \times\}$, we perform all computations modulo 23, following \citep{ye2025PoLM21}. The system prompt is shown below.

\begin{promptbox}{System Prompt}
You are an assistant that performs sequential arithmetic tasks, where all calculations must be done modulo 23.
\end{promptbox}

For the CoT data format, following \citep{ye2025PoLM21,ye2025PoLM22,zhou2025gsminfty}, we use the template
``Define [parameter] as [variable]; so [variable] [operation] = [value].''
Here, $f_i$ in \Cref{fig:main} corresponds to the operation, and $s_i$ corresponds to the value. Below, we show concrete examples of the CoT templates used in \Cref{sec:experiments}.

\begin{promptbox}{Example of Problem}
The number of hats equals 4. The number of chairs equals the number of harps minus 1. The number of harps equals the number of cats times 2. The number of blocks equals 3. The number of nautiluses equals 1. The number of scarves equals the number of spruces minus 4. The number of celery equals the number of whelks times 3. The number of boots equals the number of chairs times 2. The number of spruces equals the number of onions minus 1. The number of whelks equals 1. The number of batons equals 2. The number of maples equals 1. The number of onions equals the number of boots times 3. The number of opals equals 3. The number of needles equals 4. The number of cats equals the number of celery times 4. What is the number of scarves?
\end{promptbox}

\paragraph{Explicit CoT.}

\begin{promptbox}{Example of Explicit CoT}
Define whelks as JK; so JK = 1. Define celery as FP; so FP = JK * 3 = 1 * 3 = 3. Define cats as VK; so VK = FP * 4 = 3 * 4 = 12. Define harps as RQ; so RQ = VK * 2 = 12 * 2 = 1. Define chairs as DP; so DP = RQ - 1 = 1 - 1 = 0. Define boots as DE; so DE = DP * 2 = 0 * 2 = 0. Define onions as MP; so MP = DE * 3 = 0 * 3 = 0. Define spruces as UA; so UA = MP - 1 = 0 - 1 = 22. Define scarves as TH; so TH = UA - 4 = 22 - 4 = 18. Answer: 18.
\end{promptbox}

\paragraph{Composed CoT.}

\begin{promptbox}{Example of Composed CoT ($\texttt{g}=2$)}
Define whelks as JK; so JK = 1. Define cats as VK; so VK = JK * 3 * 4 = 1 * 3 * 4 = 12. Define chairs as DP; so DP = VK * 2 - 1 = 12 * 2 - 1 = 0. Define onions as MP; so MP = DP * 2 * 3 = 0 * 2 * 3 = 0. Define scarves as TH; so TH = MP - 1 - 4 = 0 - 1 - 4 = 18. Answer: 18.
\end{promptbox}

\begin{promptbox}{Example of Composed CoT ($\texttt{g}=4$)}
Define whelks as JK; so JK = 1. Define chairs as DP; so DP = JK * 3 * 4 * 2 - 1 = 1 * 3 * 4 * 2 - 1 = 0. Define scarves as TH; so TH = DP * 2 * 3 - 1 - 4 = 0 * 2 * 3 - 1 - 4 = 18. Answer: 18.
\end{promptbox}

\begin{promptbox}{Example of Composed CoT (g=8)}
Define whelks as JK; so JK = 1. Define scarves as TH; so TH = (JK * 3 * 4 * 2 - 1) * 2 * 3 - 1 - 4 = (1 * 3 * 4 * 2 - 1) * 2 * 3 - 1 - 4 = 18. Answer: 18.
\end{promptbox}

\paragraph{Implicit CoT.}

\begin{promptbox}{Example of Implicit CoT ($\texttt{g}=2$)}
Define whelks as JK; so JK = 1. Define cats as VK; so VK = 3 * 4 = 12. Define chairs as DP; so DP = 1 - 1 = 0. Define onions as MP; so MP = 0 * 3 = 0. Define scarves as TH; so TH = 22 - 4 = 18. Answer: 18.
\end{promptbox}

\begin{promptbox}{Example of Implicit CoT ($\texttt{g}=4$)}
Define whelks as JK; so JK = 1. Define chairs as DP; so DP = 1 - 1 = 0. Define scarves as TH; so TH = 22 - 4 = 18. Answer: 18.
\end{promptbox}

\begin{promptbox}{Example of Implicit CoT (g=8)}
Define whelks as JK; so JK = 1. Define scarves as TH; so TH = 22 - 4 = 18. Answer: 18.
\end{promptbox}

\paragraph{CoT Order.}
Backward CoT and Hierarchical CoT realize CoT by sequentially updating variables. Although replacing modular arithmetic over modulo 23 with polynomial expansion complicates symbolic algebraic manipulation, this difficulty is specific to the present sequential setting. Future work may extend the analysis to settings with branching structures.

\begin{promptbox}{Example of Forward CoT}
Define whelks as JK; so JK = 1. Define celery as FP; so FP = JK * 3 = 1 * 3 = 3. Define cats as VK; so VK = FP * 4 = 3 * 4 = 12. Define harps as RQ; so RQ = VK * 2 = 12 * 2 = 1. Define chairs as DP; so DP = RQ - 1 = 1 - 1 = 0. Define boots as DE; so DE = DP * 2 = 0 * 2 = 0. Define onions as MP; so MP = DE * 3 = 0 * 3 = 0. Define spruces as UA; so UA = MP - 1 = 0 - 1 = 22. Define scarves as TH; so TH = UA - 4 = 22 - 4 = 18. Answer: 18.
\end{promptbox}

\begin{promptbox}{Example of Backward CoT}
Define whelks as PE; so PE = 1. Define scarves as SU; so SU = LY - 4. Define spruces as LY; so SU = LY - 4 = (PY - 1) - 4 = PY - 5. Define onions as PY; so SU = PY - 5 = (TD * 3) - 5 = 3TD - 5. Define boots as TD; so SU = 3TD - 5 = 3(UD * 2) - 5 = 6UD - 5. Define chairs as UD; so SU = 6UD - 5 = 6(KF - 1) - 5 = 6KF - 11. Define harps as KF; so SU = 6KF - 11 = 6(BP * 2) - 11 = 12BP - 11. Define cats as BP; so SU = 12BP - 11 = 12(QO * 4) - 11 = 2QO - 11. Define celery as QO; so SU = 2QO - 11 = 2(PE * 3) - 11 = 6PE - 11 = 6(1) - 11 = 18. Answer: 18.
\end{promptbox}

\begin{promptbox}{Example of Hierarchical CoT}
Define whelks as PE; so PE = 1. Define celery as QO; so QO = 1 * 3 = 3. Define cats as BP; so BP = QO * 4 = 3 * 4 = 12. Define harps as KF; so KF = BP * 2 = 2BP. Define chairs as UD; so UD = KF - 1 = 2BP - 1. Define boots as TD; so TD = UD * 2 = 2UD. Define onions as PY; so PY = TD * 3 = 2UD * 3 = 6UD. Define spruces as LY; so LY = PY - 1. Define scarves as SU; so SU = LY - 4 = (PY - 1) - 4 = PY - 5. UD = 2BP - 1 = 2 * 12 - 1 = 0. SU = PY - 5 = 6UD - 5. SU = 6UD - 5 = 6 * 0 - 5 = 18. Answer: 18.
\end{promptbox}

\section{Experimental setup}\label{app:training}

\paragraph{Model}
We use Qwen2.5 \citep{yang2025qwen25} and Llama 3 \citep{grattafiori2024llama3}, both of which are decoder-only dense Transformers. We use these models in accordance with their respective licenses and terms of use: Qwen2.5 is released under the Apache 2.0 license, and Llama 3 is released under the Llama 3 Community License.

\paragraph{Training}
Unless noted otherwise, all experiments are conducted on NVIDIA H100 and GH200 GPUs. We use a batch size of 48 for each optimization step. SFT is performed using LLaMA-Factory \citep{zheng2024llamafactory}, with the configuration shown in \Cref{tab:sft-config}. For RLVR, we use GRPO \citep{shao2024deepseekmath} implemented in the verl framework \citep{sheng2024hybridflow}, with hyperparameters listed in \Cref{tab:rlvr-config}.

\begin{table}[h]
\centering
\caption{SFT Configuration.}
\label{tab:sft-config}
\begin{tabular}{ll}
\toprule
\textbf{Component} & \textbf{Setting} \\
\midrule
Effective batch size & 48 \\
Optimizer & AdamW \\
Learning rate & $2.0 \times 10^{-5}$ \\
Weight decay & 0.1 \\
Max gradient norm & 1.0 \\
Scheduler & Cosine \\
Warmup ratio & 0.05 \\
Minimum learning rate & $3.0 \times 10^{-6}$ \\
Mixed precision & bfloat16 \\
\bottomrule
\end{tabular}
\end{table}

\begin{table}[h]
\centering
\caption{RLVR Configuration.}
\label{tab:rlvr-config}
\begin{tabular}{ll}
\toprule
\textbf{Component} & \textbf{Setting} \\
\midrule
Rollouts & 8 \\
Sampling temperature & 1.0 \\
Top-$p$ & 1.0 \\
Maximum response length & 2048 \\
Training batch size & 256 \\
Actor learning rate & $1.0 \times 10^{-6}$ \\
Weight decay & 0.01 \\
Maximum gradient norm & 1.0 \\
KL coefficient & 0.001 \\
\bottomrule
\end{tabular}
\end{table}

\paragraph{Evaluation.}

We evaluate each model on $\texttt{op}=32,40,48,\ldots,104$ tasks, where each task has 336 samples. Evaluation is performed with greedy decoding ($\text{temperature}=0$ and $\text{top_p}=1$). The maximum number of new tokens is set according to the $\texttt{op}$ range: 4096 for $\texttt{op}=25$–44, 8192 for $\texttt{op}=45$–64, 12288 for $\texttt{op}=65$–84, and 16384 for $\texttt{op}=85$–104, which provide enough generation budget for correct solutions.

To assess the reliability of our numerical results, we conduct additional evaluations with stochastic decoding. Specifically, we report the mean and standard deviation of pass@1 over five sampling runs with a temperature of 0.6 and top-$p$ of 0.95. As shown in \Cref{tab:sampling_variance}, the standard deviations remain small across models, CoT types, and numbers of \texttt{op}s, indicating that the observed performance differences are robust to sampling variability. In particular, the variance remains sufficiently small even though each \texttt{op} setting contains only 336 evaluation samples.

\begin{table*}[t]
\centering
\small
\begin{tabular}{llccc}
\toprule
\textbf{Model} & \textbf{CoT Types} &
\multicolumn{3}{c}{\textbf{Number of \texttt{op}}} \\
\cmidrule(lr){3-5}
& & \textbf{32} & \textbf{64} & \textbf{96} \\
\midrule
Qwen2.5-3B
& Explicit & $99.82 \pm 0.15$ & $96.61 \pm 0.48$ & $94.46 \pm 0.61$ \\
& Composed & $99.94 \pm 0.12$ & $97.68 \pm 0.22$ & $92.32 \pm 0.48$ \\
& Implicit & $99.94 \pm 0.12$ & $97.62 \pm 0.33$ & $87.56 \pm 0.61$ \\
\midrule
Llama-3.2-3B-Instruct
& Explicit & $100.00 \pm 0.00$ & $99.58 \pm 0.24$ & $79.11 \pm 1.10$ \\
& Composed & $100.00 \pm 0.00$ & $98.15 \pm 0.12$ & $90.54 \pm 0.79$ \\
& Implicit & $98.99 \pm 0.15$ & $96.49 \pm 0.35$ & $81.61 \pm 1.10$ \\
\bottomrule
\end{tabular}
\caption{\textbf{Reliability of evaluation results under stochastic decoding.}
Mean and standard deviation of pass@1 (\%) over five sampling runs with temperature $0.6$ and top-$p$ $0.95$.}
\label{tab:sampling_variance}
\end{table*}

\paragraph{Prompt Template.}
We use the following prompt templates for Qwen2.5 models and Llama-3 models, respectively.

\begin{promptbox}{Prompt Template for Qwen2.5 Models}
<|im_start|>system
{system}<|im_end|>
<|im_start|>user
{user}<|im_end|>
<|im_start|>assistant
{assistant}<|im_end|>
\end{promptbox}

\begin{promptbox}{Prompt Template for Llama3 Models}
<|begin_of_text|><|start_header_id|>system<|end_header_id|>

{system}<|eot_id|><|start_header_id|>user<|end_header_id|>

{user}<|eot_id|><|start_header_id|>assistant<|end_header_id|>

{assistant}<|eot_id|>
\end{promptbox}

\section{Detailed Results}
In this section, we report the detailed results for \Cref{sec:experiments}.

\subsection{SFT Results on Different CoT Datasets}\label{app:data_granularity}

\begin{figure*}[t]
  \centering
  \includegraphics[width=\linewidth]{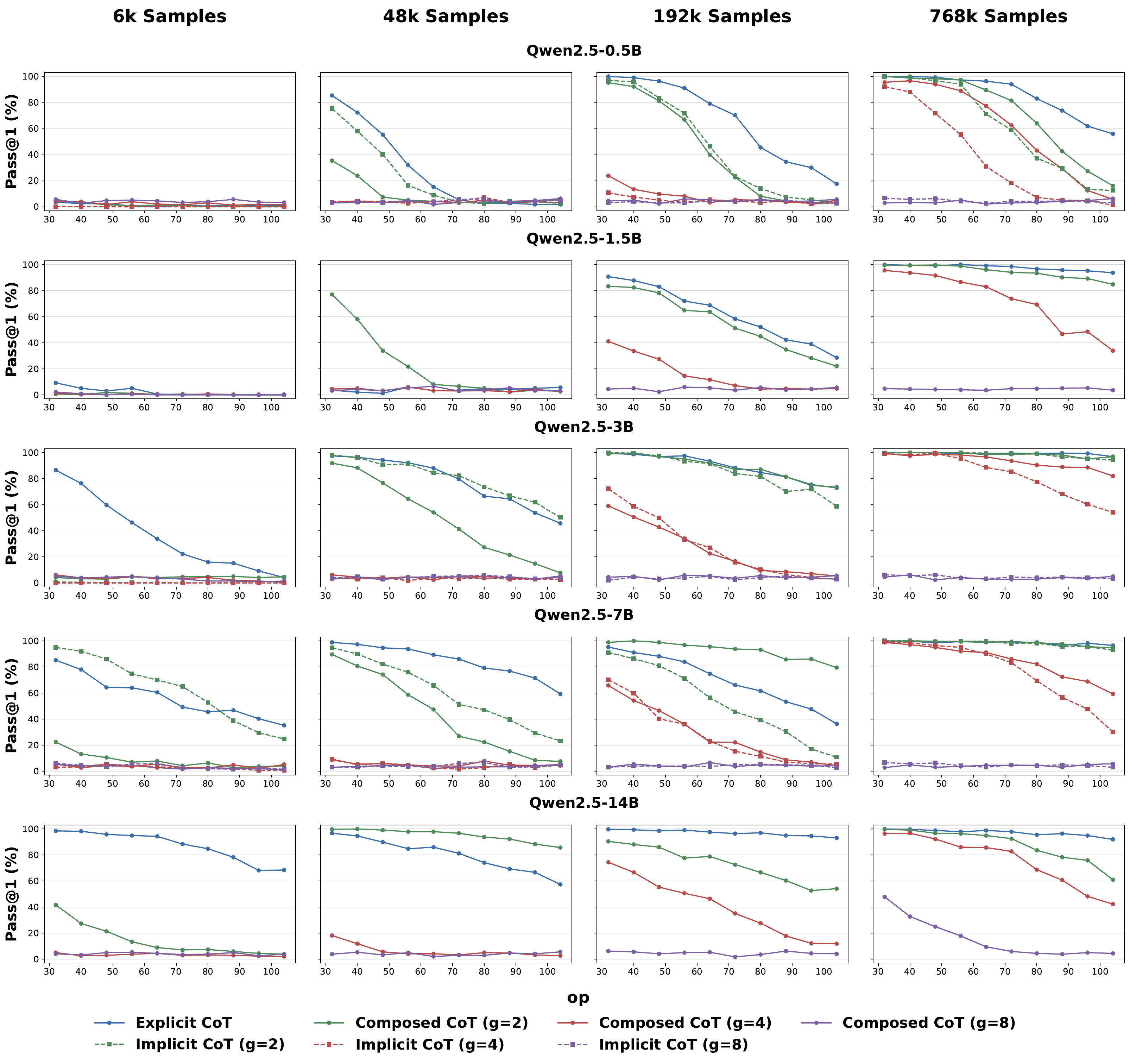}
  \caption{\textbf{Evaluation Results of Qwen2.5 Models.} Evaluation results on $\texttt{op}=32,40,48,\ldots,96,104$ tasks for checkpoints after SFT for one epoch with 6k, 48k, 192k, and 768k samples for each CoT type at $\texttt{op}=8,16,24$ tasks.}
  \label{fig:eval_qwen}
\end{figure*}

\begin{figure*}[t]
  \centering
  \includegraphics[width=\linewidth]{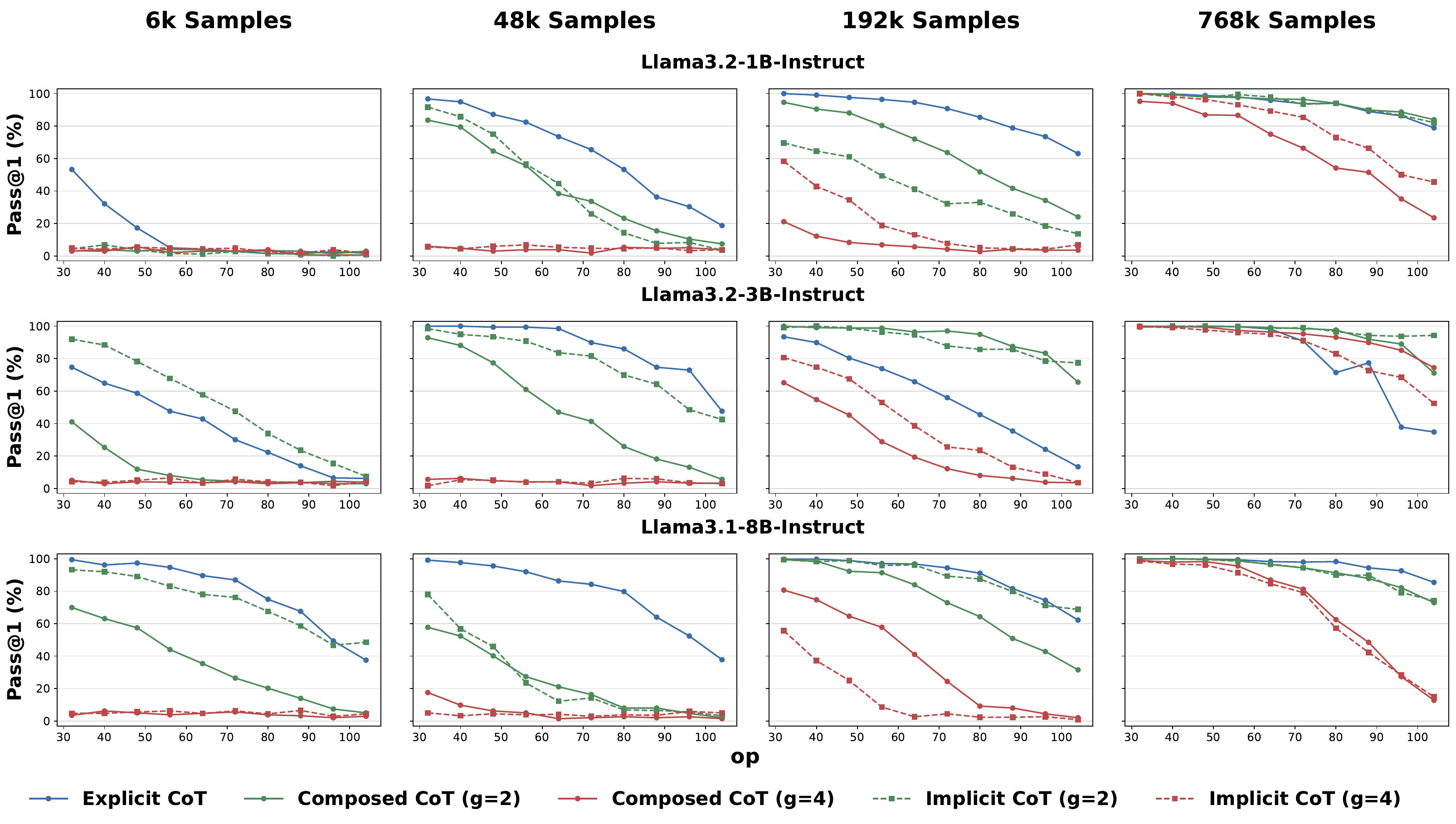}
  \caption{\textbf{Evaluation Results of Llama-3 Models.} Evaluation results on $\texttt{op}=32,40,48,\ldots,96,104$ tasks for checkpoints after SFT for one epoch with 6k, 48k, 192k, and 768k samples for each CoT type at $\texttt{op}=8,16,24$ tasks.}
  \label{fig:eval_llama}
\end{figure*}

For Qwen2.5-0.5B, 1.5B, 3B, 7B, and 14B, we use Explicit CoT, Composed CoT with $\texttt{g}=2,4,8$, and Implicit CoT with $\texttt{g}=2,4,8$. For Llama-3.2-1B-Instruct, Llama-3.2-3B-Instruct, and Llama-3.1-8B-Instruct, we use Explicit CoT, Composed CoT with $\texttt{g}=2,4$ , and Implicit CoT with $\texttt{g}=2,4$. Under each setting, we perform SFT on $\texttt{op}=8,16,24$ ($\equiv 0 \pmod{8}$) tasks, and evaluate on $\texttt{op}=32,40,48,\ldots,96,104$ ($\equiv 0 \pmod{8}$) tasks. \Cref{fig:eval_qwen} and \Cref{fig:eval_llama} show the evaluation results at checkpoints obtained after one epoch with 6K, 48K, 192K, and 768K training samples.

\paragraph{Math Experiments}
We performed full-parameter SFT of Qwen2.5-7B on 5586 Level 3--5 problems from the Hendrycks-MATH training set. The reasoning traces were represented using Explicit, Composed, or Implicit CoT, with the Composed and Implicit traces generated by Qwen3.5-Flash. See the prompts of Qwen3.5-Flash below. Each model was trained for two epochs with an effective batch size of 64, a learning rate of $2 \times 10^{-5}$, a cosine learning rate schedule, a warmup ratio of 0.03, and BF16 precision. We evaluated checkpoints after 44, 88, and 176 optimization steps, corresponding to 0.5, 1, and 2 epochs, respectively. Evaluation was conducted on GSM8K, MATH500, and AMC23 using greedy decoding with a maximum generation length of 4096 tokens, and performance was reported as pass@1 (\%).

\begin{promptbox}{System Prompt for Qwen3.5-Flash}
You rewrite correct mathematical solutions. Preserve the problem facts, logical validity, notation, and exact final answer. Never solve a different problem.
\end{promptbox}

\begin{promptbox}{Prompt for Rewriting Solutions into Composed CoT Using Qwen3.5-Flash}
Rewrite the Original Solution in Composed CoT style.

Rewrite rule: Compose adjacent algebraic or logical steps into larger steps. Keep the main derivation visible, but remove routine expansions, repeated explanations, and minor intermediate values.

Compression requirement: target approximately 1/3 to 1/6 of the Original Solution's length. Prefer the shortest version that remains correct and self-contained; do not merely paraphrase sentence by sentence.

Correctness requirement: preserve the reasoning outcome and exact final answer. Do not introduce a new method unless needed to achieve the compression.

Return only the rewritten solution as plain text. Do not add commentary, labels, XML/HTML tags, <think> tags, or markdown fences. Preserve the final answer, including its LaTeX and boxed form when present.

Problem:

{problem}

Original Solution:

{solution}
\end{promptbox}

\begin{promptbox}{Prompt for Rewriting Solutions into Implicit CoT Using Qwen3.5-Flash}
Rewrite the Original Solution in Implicit CoT style.

Rewrite rule: Make routine algebra, arithmetic, and obvious local deductions implicit. Keep only the key setup, decisive transitions, boundary results, and final answer needed for a mathematically competent reader to verify the solution.

Compression requirement: target approximately 1/3 to 1/6 of the Original Solution's length. Prefer the shortest version that remains correct and self-contained; do not merely paraphrase sentence by sentence.

Correctness requirement: preserve the reasoning outcome and exact final answer. Do not introduce a new method unless needed to achieve the compression.

Return only the rewritten solution as plain text. Do not add commentary, labels, XML/HTML tags, <think> tags, or markdown fences. Preserve the final answer, including its LaTeX and boxed form when present.

Problem:

{problem}

Original Solution:

{solution}
\end{promptbox}

\subsection{SFT Results on Different Number of Epochs.}\label{app:data_diversity}
\begin{figure*}[t]
  \centering
  \includegraphics[width=\linewidth]{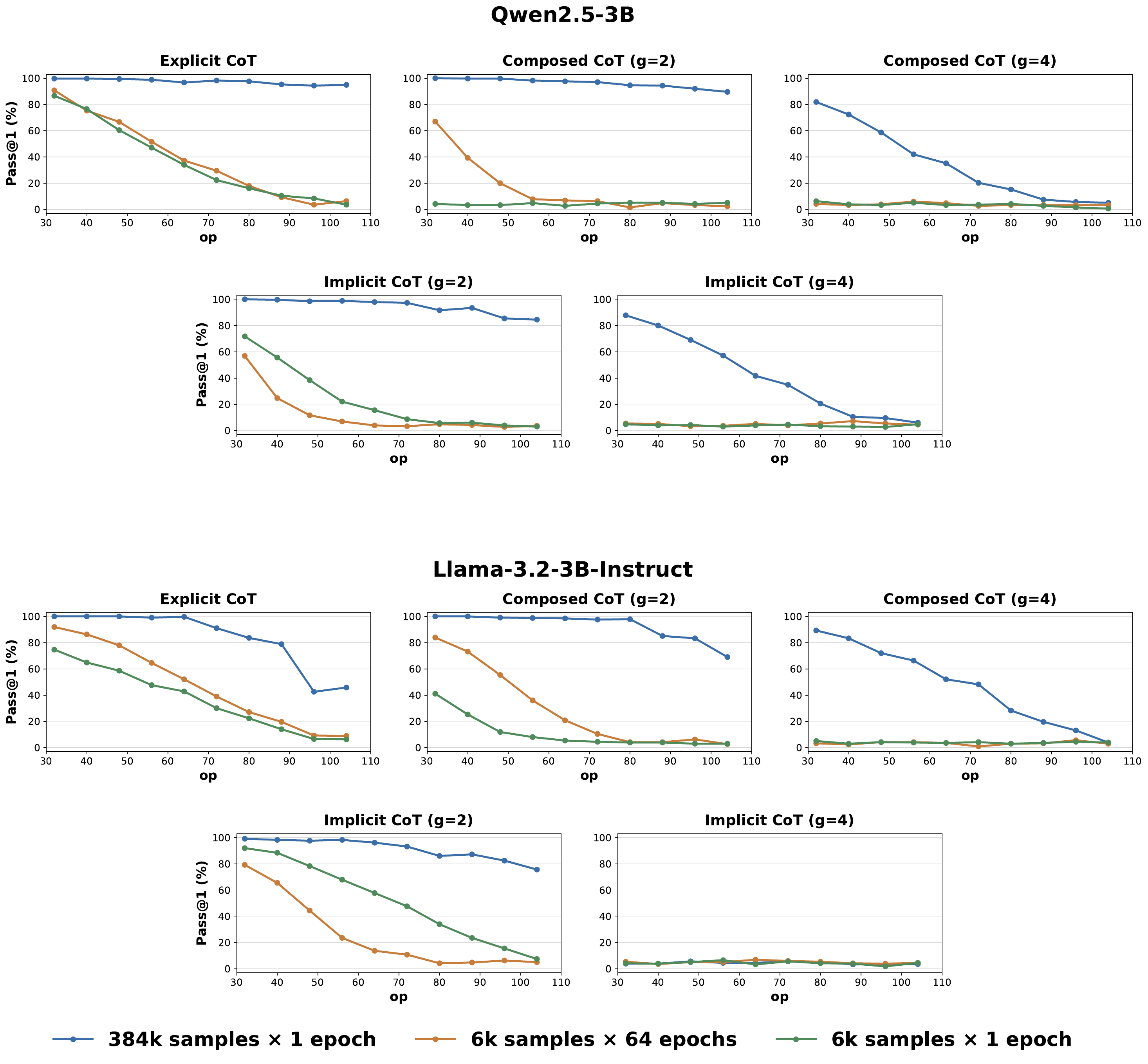}
  \caption{\textbf{Evaluation Results of Different Number of Epochs.} Evaluation results on $\texttt{op}=32,40,48,\ldots,96,104$ for checkpoints after SFT for one epoch with 384k samples for one epoch, 6k samples for 64 epochs, and 6k samples for one epoch for each CoT type at $\texttt{op}=8,16,24$ tasks.}
  \label{fig:eval_data_diversity}
\end{figure*}

For Qwen2.5-3B and Llama-3.2-3B-Instruct, we consider Explicit CoT, Composed CoT with $\texttt{g}=2,4$, and Implicit CoT with $\texttt{g}=2,4$. We perform SFT on $\texttt{op}=8,16,24$ ($\equiv 0 \pmod{8}$) tasks under three training regimes: one epoch on 384K samples, 64 epochs on 6K samples, and one epoch on 6K samples. \Cref{fig:eval_data_diversity} shows the evaluation results on $\texttt{op}=32,40,48,\ldots,96,104$ ($\equiv 0 \pmod{8}$ tasks).

\begin{figure*}[t]
  \centering
  \includegraphics[width=\linewidth]{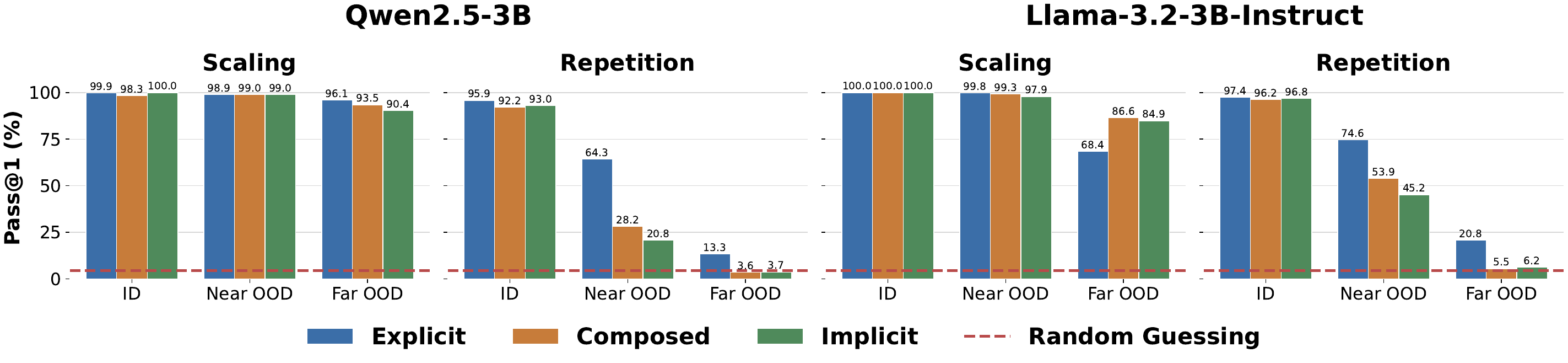}
  \caption{\textbf{Sequence-length generalization under scaling and data repetition.} Performance (\%) of Qwen2.5-3B and Llama-3.2-3B-Instruct on ID ($\texttt{op}=8,16,24$), Near OOD ($\texttt{op}=32,40,\ldots,64$), and Far OOD ($\texttt{op}=72,80,\ldots,104$) tasks. Data scaling uses 384k samples for 1 epoch, whereas data repetition uses 6k samples for 64 epochs. 
}
  \label{fig:cot_generalization}
\end{figure*}

\Cref{fig:cot_generalization} shows that, under data repetition, performance degrades increasingly as the OOD distance grows, particularly for Implicit CoT, even when ID performance remains high. Here, ID denotes $\texttt{op}=8,16,24$, Near OOD denotes $\texttt{op}=32,40,\ldots,64$, and Far OOD denotes $\texttt{op}=72,80,\ldots,104$. This suggests that the model can fit shorter ID tasks but struggles to generalize to longer OOD sequences.

\subsection{RLVR Results}\label{app:rlvr}
\begin{figure*}[t]
  \centering
  \includegraphics[width=\linewidth]{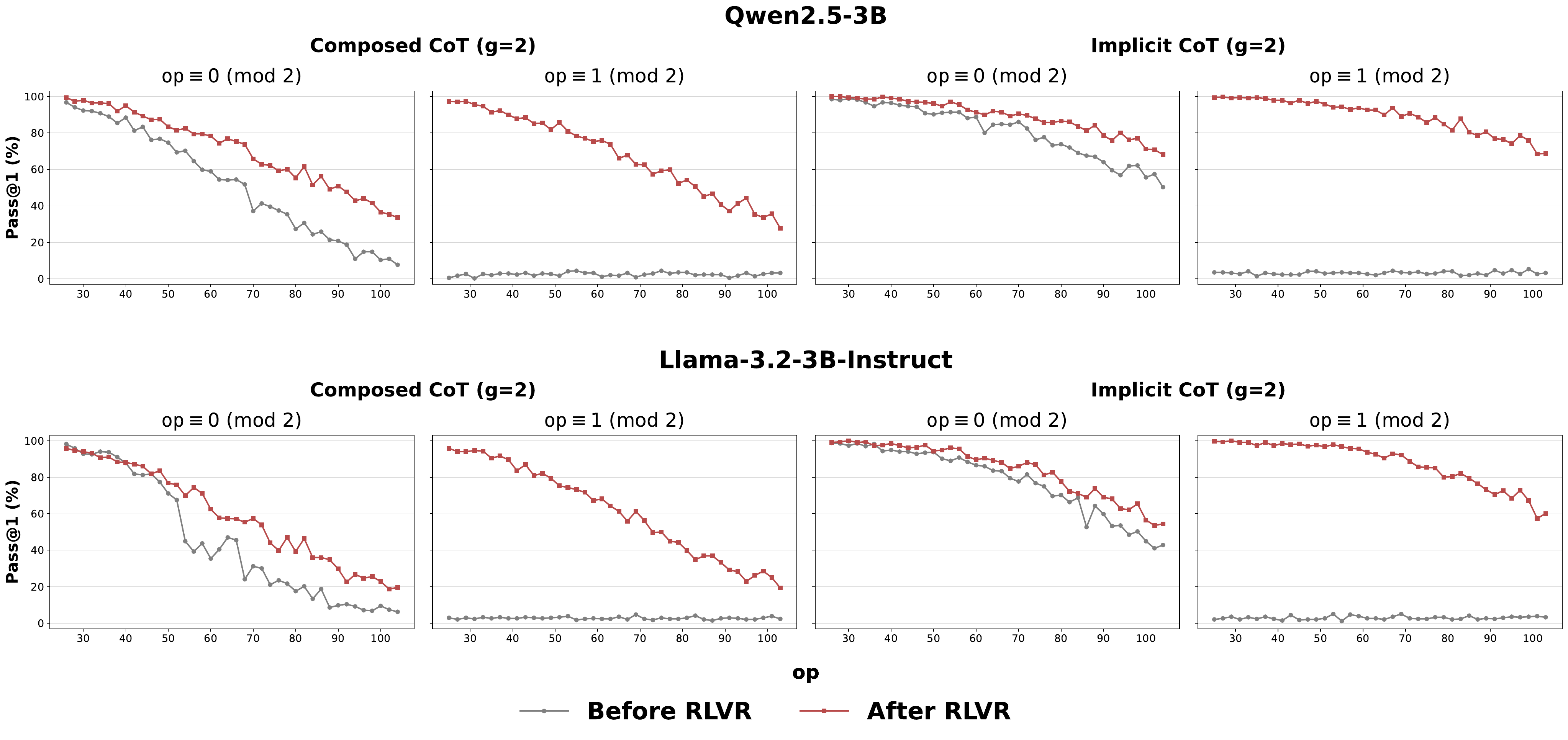}
  \caption{\textbf{Evaluation Results of Before and After RLVR on odd \texttt{op} tasks.} Evaluation results on $\texttt{op}=25,27,29,\ldots,101,103$ tasks for checkpoints after RLVR on $\texttt{op}=9,11,13,15$ tasks for each CoT type.}
  \label{fig:eval_rlvr}
\end{figure*}

For Qwen2.5-3B and Llama-3.2-3B-Instruct, we start from checkpoints obtained by one-epoch SFT on 24K samples with $\texttt{op}=8,16,24$, using Explicit CoT, Composed CoT with $\texttt{g}=2$, and Implicit CoT with $\texttt{g}=2$. We then perform RLVR with GRPO on $\texttt{op}=9,11,13,15$ and $\texttt{op}=10,12,14$ tasks using the configuration in \Cref{tab:rlvr-config}. \Cref{fig:eval_rlvr} shows the evaluation results on $\texttt{op}=32,40,48,\ldots,96,104$ ( $\texttt{op} \equiv 0 \pmod{8}$) tasks.

\begin{figure*}[t]
  \centering
  \begin{subfigure}{\linewidth}
    \centering
    \includegraphics[width=0.95\linewidth]{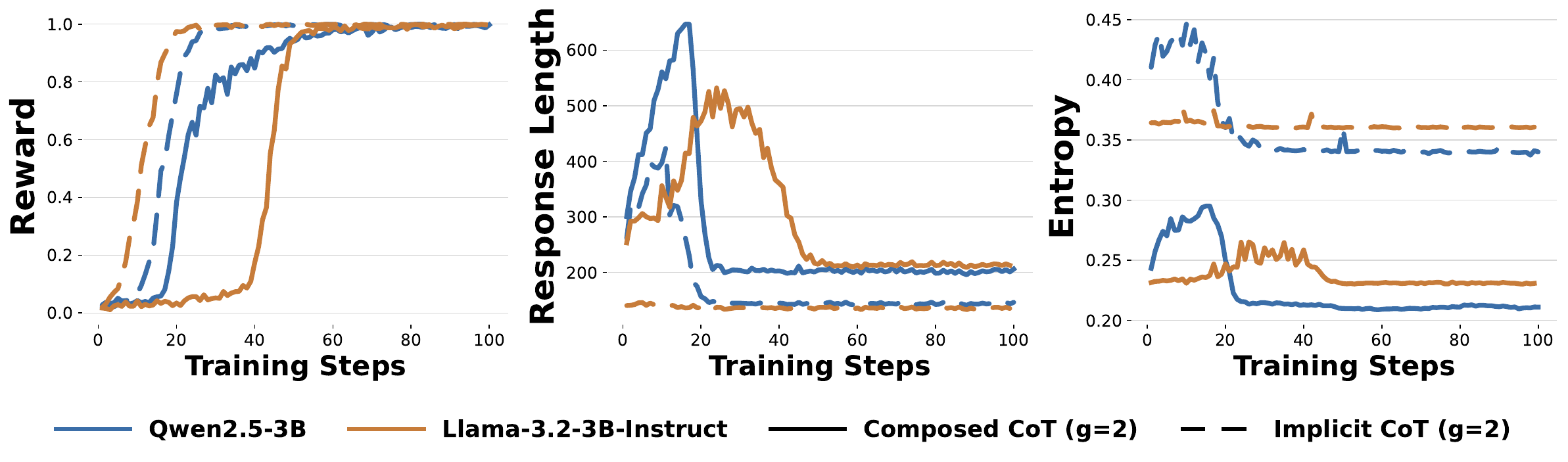}
    \caption{Odd \texttt{op} tasks}
    \label{fig:eval_rlvr_odd}
  \end{subfigure}
  \begin{subfigure}{\linewidth}
    \centering
    \includegraphics[width=0.95\linewidth]{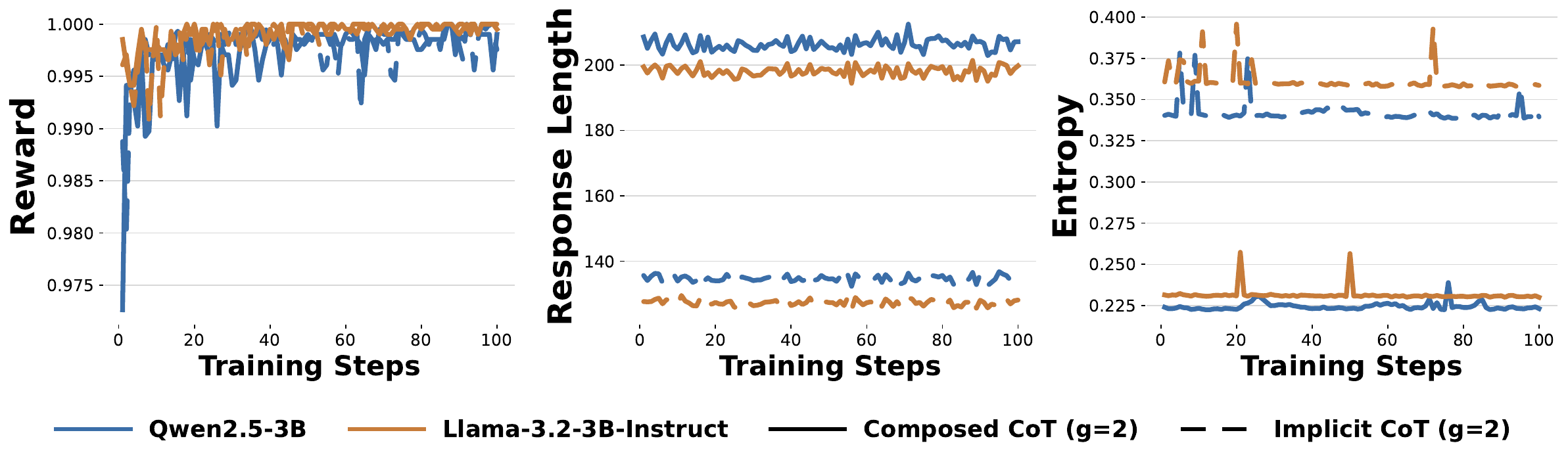}
    \caption{Even \texttt{op} tasks}
    \label{fig:eval_rlvr_even}
  \end{subfigure}
  \caption{\textbf{Evaluation results before and after RLVR on odd and even \texttt{op} tasks.} Training dynamics of the mean reward, mean rollout response length, and mean token entropy at each steps. Odd \texttt{op} task are $\texttt{op}=9,11,13,15$ and even \texttt{op} task are $\texttt{op}=10,12,14$.}
  \label{fig:eval_rlvr_odd_even}
\end{figure*}

\Cref{fig:eval_rlvr_odd_even} shows the training dynamics of RLVR. In terms of response length, $\texttt{op}=10,12,14$ exhibit nearly constant lengths throughout training. By contrast, for $\texttt{op}=9,11,13,15$, the response length starts to increase when the reward begins to improve, and subsequently converges.
This indicates that the model first decomposes chunks obtained from exploration with $\texttt{g}=2$ into smaller chunks with $\texttt{g}=1$. Eventually, it performs CoT with $\texttt{g}=2$ while processing fractions with $\texttt{g}=1$.

\begin{figure*}[t]
  \centering
\includegraphics[width=0.95\linewidth]{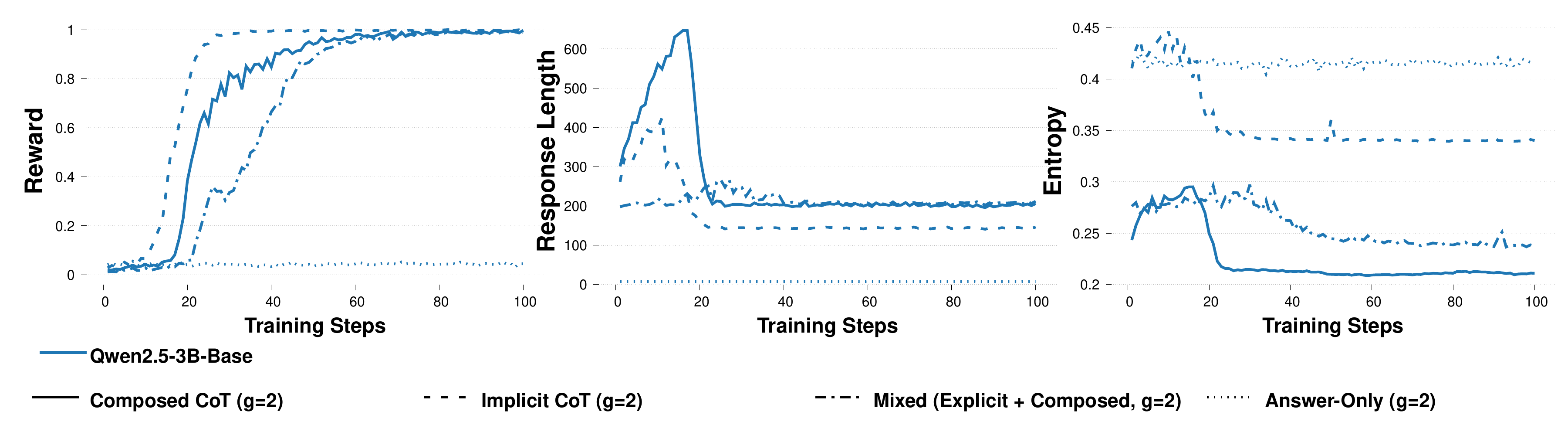}
  \caption{\textbf{Evaluation results for the RLVR baseline ablations.} Training dynamics of the mean reward, mean rollout response length, and mean token entropy at each steps. Models are trained by odd \texttt{op} tasks where $\texttt{op}=9,11,13,15$.}
  \label{fig:rlvr_baseline_ablations}
\end{figure*}

Furthermore, \Cref{fig:rlvr_baseline_ablations} shows RLVR training dynamics of SFT with Explicit CoT, Composed CoT, mixed data consisting of equal proportions of Explicit CoT and Composed CoT (Mixed), and answer-only data without reasoning traces (Answer-only). The compression granularity of all Compressed CoT data are $\texttt{g}=2$

This shows that Explicit CoT + RLVR and Mixed SFT + RLVR improve on the odd tasks, as they are already exposed to g=1 steps during SFT. In contrast, Composed CoT + RLVR achieves reward gains on odd tasks that are unseen during SFT. Moreover, Answer-only SFT + RLVR shows no reward improvement, indicating that CoT remains essential even when compressed. These results suggest that scratchpads play a crucial role in enabling Transformer-based LLMs to solve sequential problems, and that interaction with the environment during RLVR allows the model to recover the decomposition in a compressed scratchpad.

Below are reasoning samples after RLVR under each setting. When RLVR is applied to checkpoints after SFT on Composed CoT, the model continues reasoning with $\texttt{g}=2$, Qwen2.5-3B outputs with $\texttt{g}=1$ at the end, and Llama-3.2-3B-Instruct inserts a dummy $\texttt{* 1}$ at the end. When RLVR is applied to checkpoints after SFT on Implicit CoT, the model outputs only the operation following $\texttt{g}=2$, and the final step indicates that it is handling a fraction operation ($\texttt{g}=1$).

\begin{promptbox}{Qwen2.5-3B, Composed CoT ($\texttt{g}=2$)}
Define whelks as YH; so YH = 3. Define cats as YO; so YO = YH + 1 + 4 = 3 + 1 + 4 = 8. Define chairs as ZD; so ZD = YO * 4 * 2 = 8 * 4 * 2 = 18. Define onions as YL; so YL = ZD * 3 - 2 = 18 * 3 - 2 = 6. Define scarves as YP; so YP = YL * 3 * 2 = 6 * 3 * 2 = 13. Define batons as ZQ; so ZQ = YP + 2 + 2 = 13 + 2 + 2 = 17. Define hats as ZT; so ZT = ZQ * 3 + 1 = 17 * 3 + 1 = 6. Define maples as YX; so YX = ZT - 4 + 3 = 6 - 4 + 3 = 5. Define violets as YO; so YO = YX * 3 * 2 = 5 * 3 * 2 = 7. Define flutes as YL; so YL = YO * 3 * 2 = 7 * 3 * 2 = 19. Define markers as YH; so YH = YL * 4 * 3 = 19 * 4 * 3 = 21. Define aspens as YL; so YL = (YH - 1) * 3 = (21 - 1) * 3 = 14. Define wolves as YL; so YL = YL - 2 - 2 = 14 - 2 - 2 = 10. Define YL + 2 = 10 + 2 = 12. Answer: 12.
\end{promptbox}

\begin{promptbox}{Qwen2.5-3B, Implicit CoT ($\texttt{g}=2$)}
Define whelks as ZD; so ZD = 3. Define cats as ZF; so ZF = 4 + 4 = 8. Define chairs as ZU; so ZU = 9 * 2 = 18. Define onions as ZL; so ZL = 8 - 2 = 6. Define scarves as ZC; so ZC = 18 * 2 = 13. Define batons as ZY; so ZY = 15 + 2 = 17. Define hats as ZN; so ZN = 5 + 1 = 6. Define maples as ZD; so ZD = 2 + 3 = 5. Define violets as ZF; so ZF = 15 * 2 = 7. Define flutes as ZD; so ZD = 21 * 2 = 19. Define markers as ZD; so ZD = 7 * 3 = 21. Define aspens as ZD; so ZD = 20 * 3 = 14. Define wolves as ZD; so ZD = 12 - 2 = 10. Define tokens as ZD; so ZD = 10 + 2 = 12. Answer: 12.

\end{promptbox}

\begin{promptbox}{Llama3.2-3B-Instruct, Composed CoT ($\texttt{g}=2$)}
Define whelks as ZD; so ZD = 3. Define cats as ZT; so ZT = ZD + 1 + 4 = 3 + 1 + 4 = 8. Define chairs as ZQ; so ZQ = ZT * 4 * 2 = 8 * 4 * 2 = 18. Define onions as ZD; so ZD = ZQ * 3 - 2 = 18 * 3 - 2 = 6. Define scarves as ZO; so ZO = ZD * 3 * 2 = 6 * 3 * 2 = 13. Define batons as ZD; so ZD = ZO + 2 + 2 = 13 + 2 + 2 = 17. Define hats as ZT; so ZT = ZD * 3 + 1 = 17 * 3 + 1 = 6. Define maples as ZD; so ZD = ZT - 4 + 3 = 6 - 4 + 3 = 5. Define violets as ZT; so ZT = ZD * 3 * 2 = 5 * 3 * 2 = 7. Define flutes as ZD; so ZD = ZT * 3 * 2 = 7 * 3 * 2 = 19. Define markers as ZT; so ZT = ZD * 4 * 3 = 19 * 4 * 3 = 21. Define aspens as ZD; so ZD = (ZT - 1) * 3 = (21 - 1) * 3 = 14. Define wolves as ZT; so ZT = ZD - 2 - 2 = 14 - 2 - 2 = 10. Define tokens as ZD; so ZD = ZT * 1 + 2 = 10 * 1 + 2 = 12. Answer: 12.
\end{promptbox}

\begin{promptbox}{Llama3.2-3B-Instruct, Implicit CoT ($\texttt{g}=2$)}
Define whelks as ZD; so ZD = 3. Define cats as ZC; so ZC = 4 + 4 = 8. Define chairs as ZK; so ZK = 9 * 2 = 18. Define onions as ZU; so ZU = 8 - 2 = 6. Define scarves as ZC; so ZC = 18 * 2 = 13. Define batons as ZD; so ZD = 15 + 2 = 17. Define hats as ZU; so ZU = 5 + 1 = 6. Define maples as ZC; so ZC = 2 + 3 = 5. Define violets as ZD; so ZD = 15 * 2 = 7. Define flutes as ZD; so ZD = 21 * 2 = 19. Define markers as ZD; so ZD = 7 * 3 = 21. Define aspens as ZD; so ZD = 20 * 3 = 14. Define wolves as ZD; so ZD = 12 - 2 = 10. Define tokens as ZD; so ZD = 10 + 2 = 12. Answer: 12.
\end{promptbox}

\subsection{SFT Results on Different CoT Orders}\label{app:cot_order}
\begin{figure*}[t]
  \centering
  \includegraphics[width=\linewidth]{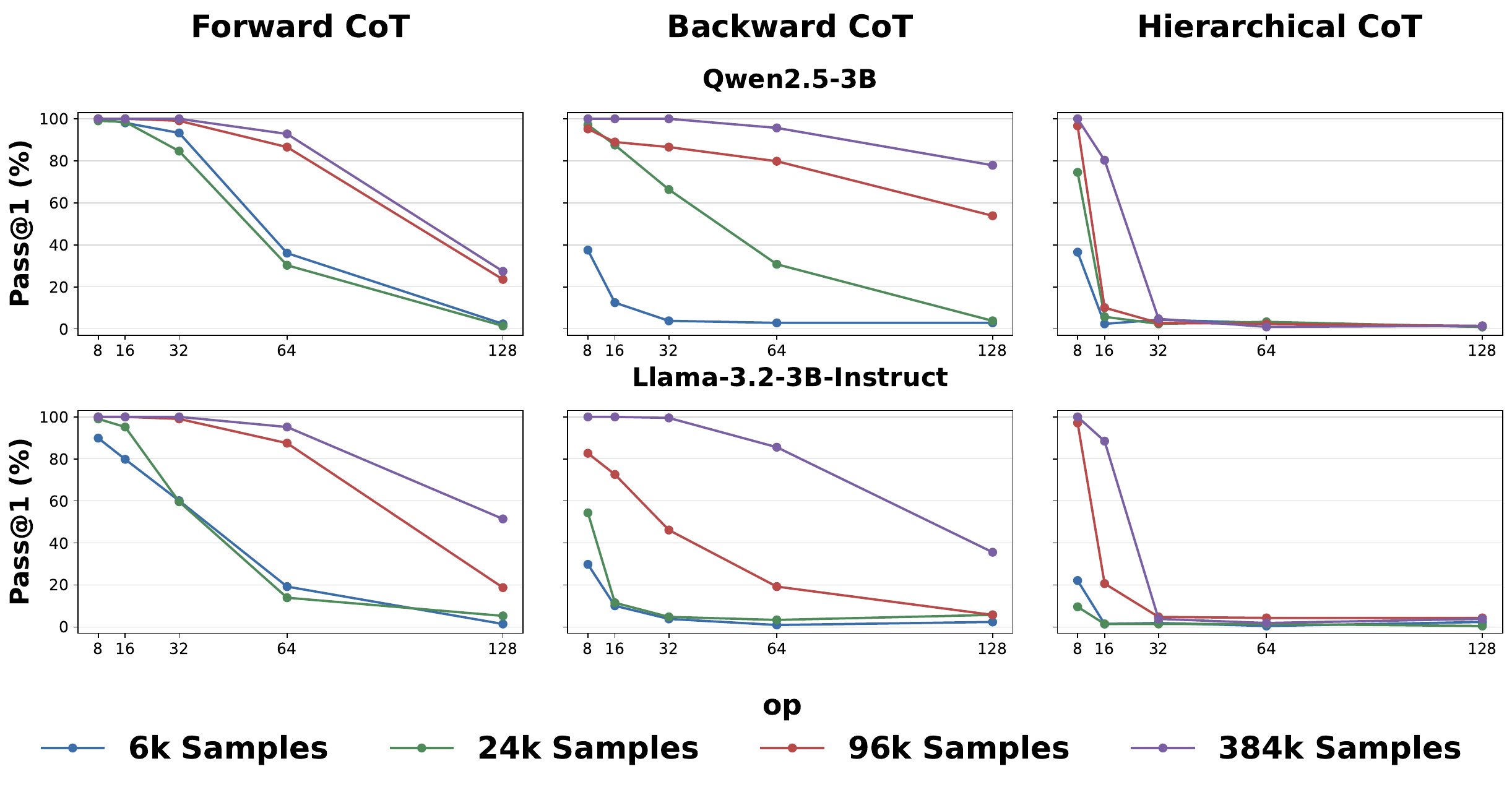}
  \caption{\textbf{Evaluation Results of Different CoT Orders.} Evaluation results on $\texttt{op}=8,16,32,64,128$ tasks for checkpoints after SFT on $\texttt{op}=8,16$ tasks for each CoT order. $\texttt{op}=8,16$ tasks are ID, and $\texttt{op}=32,64,128$ tasks are OOD.}
  \label{fig:eval_cot_order}
\end{figure*}

For Qwen2.5-3B and Llama-3.2-3B-Instruct, we consider Forward CoT, Backward CoT, and Hierarchical CoT. We perform SFT on $\texttt{op}=8,16$ tasks ($\equiv 0 \pmod{8}$), varying the training dataset size among 6k, 24k, 96k, and 384k. \Cref{fig:eval_cot_order} shows the evaluation results on $\texttt{op}=32,64,128$ ($\equiv 0 \pmod{8}$ tasks.

\end{document}